\documentclass[acmtog]{acmart}
\usepackage{booktabs} 

\usepackage[ruled]{algorithm2e} 

\SetAlFnt{\small}
\SetAlCapFnt{\small}
\SetAlCapNameFnt{\small}
\SetAlCapHSkip{0pt}

\acmJournal{TOG}

\usepackage{orcidlink}
\usepackage{xcolor}
\usepackage{graphicx}
\usepackage{amsmath}
\usepackage{capt-of}
\usepackage{adjustbox}
\usepackage{cuted}
\usepackage{subfigure}
\usepackage[utf8]{inputenc} 
\usepackage[T1]{fontenc}    
\usepackage{hyperref}       
\usepackage{url}            
\usepackage{booktabs}       
\usepackage{amsfonts}       
\usepackage{nicefrac}       
\usepackage{microtype}      
\usepackage{xcolor}         
\usepackage{kotex}
\usepackage{amsmath}
\usepackage{booktabs}    
\usepackage{multirow}    
\usepackage{graphicx}    
\usepackage{algorithmic} 
\usepackage{amsthm, enumitem} 
\usepackage[bottom, perpage]{footmisc} 
\usepackage{lipsum}
\usepackage{bbding}

\newcommand{\figref}[1]{Figure~\ref{#1}}
\newcommand{\tabref}[1]{Table~\ref{#1}}
\newcommand{\secref}[1]{Section~\ref{#1}}
\newcommand{\algref}[1]{Algorithm~\ref{#1}}
\newcommand{\equref}[1]{Equation~(\ref{#1})}

\newcommand{\rev}[1]{{#1}}

\begin{document}
\title{CRAFT: Constrained Reward via Attention Fine-Tuning for Subject Personalization without Composed Targets}

\author{Jihun Park}
\authornote{Work done during an internship at Baidu, Inc.}
\orcid{0009-0004-1072-1239}
\affiliation{%
  \institution{DGIST}
  \city{Daegu}
  \country{South Korea}}
\email{pjh2857@dgist.ac.kr}

\author{Kyoungmin Lee}
\orcid{0009-0008-2581-3610}
\affiliation{%
  \institution{DGIST}
  \city{Daegu}
  \country{South Korea}}
\email{kyoungmin@dgist.ac.kr}

\author{Jongmin Gim}
\orcid{0009-0005-9082-7427}
\affiliation{%
  \institution{DGIST}
  \city{Daegu}
  \country{South Korea}}
\email{jongmin4422@dgist.ac.kr}

\author{Hyeonseo Jo}
\orcid{0009-0006-1294-3411}
\affiliation{%
  \institution{DGIST}
  \city{Daegu}
  \country{South Korea}}
\email{gustj0510@dgist.ac.kr}

\author{Jaeyeul Kim}
\orcid{0000-0002-7765-4972}
\affiliation{%
  \institution{DGIST}
  \city{Daegu}
  \country{South Korea}}
\email{jykim94@dgist.ac.kr}

\author{Han Zou}
\orcid{0009-0009-1657-1573}
\affiliation{%
  \institution{Baidu, Inc.}
  \city{Shenzhen}
  \country{China}}
\email{zouhan@baidu.com}

\author{Zhenpeng Zhan}
\orcid{0009-0004-2333-3796}
\affiliation{%
  \institution{Baidu, Inc.}
  \city{Shenzhen}
  \country{China}}
\email{zhanzhenpeng01@baidu.com}

\author{Yan Zhang}
\authornote{Corresponding authors.}
\orcid{0009-0009-1041-1990}
\affiliation{%
  \institution{Baidu, Inc.}
  \city{Shenzhen}
  \country{China}}
\email{zhangyan97@baidu.com}

\author{Sunghoon Im}
\authornotemark[2]
\orcid{0000-0001-9776-8101}
\affiliation{%
  \institution{KAIST}
  \city{Daejeon}
  \country{South Korea}}
\email{im@kaist.ac.kr}

\renewcommand{\shortauthors}{Park et al.}

\begin{abstract}
Subject-driven image personalization---generating new images that preserve the identity of one or several reference subjects in novel scenes---is a foundational capability for modern visual content creation. It is currently dominated by generalized methods that fine-tune a pretrained multimodal diffusion transformer (MMDiT) on hundreds of thousands to millions of paired \emph{(reference, composed-target)} examples, where each composed target is a synthesized image of the subject in a novel scene. Producing such targets demands a costly multi-stage curation pipeline---LLM-based prompt generation, T2I-based composed-target synthesis, reference-subject extraction, VLM-based quality filtering, and correspondence labeling---and tightly couples each method to a particular target synthesizer and curation choice. We introduce \emph{CRAFT} (Constrained Reward via Attention Fine-Tuning), a single-step ReFL framework that fine-tunes a pre-trained \emph{reference-aware} MMDiT via LoRA adapters using a compact reference-only data construction---$10$K reference images and subject masks, with no composed-target supervision. CRAFT realizes a \emph{Where to look} principle: attention-level rewards align noise- and phrase-token attention with the correct reference subject, and the resulting per-subject attention masks gate a pixel-level identity reward to keep image-space supervision consistent with the learned attention routing. Applied to FLUX.2-klein-9B, CRAFT achieves state-of-the-art performance on XVerseBench \rev{while using no composed-target supervision---only $10$K reference-only samples, whereas prior generalized methods require $150$K to over $2$M composed-target pairs}. The same recipe transfers to other reference-aware backbones, consistently improving performance. Project page: \url{https://jihun999.github.io/projects/CRAFT/}.
\end{abstract}

%
%

%


\maketitle

\section{Introduction}
\label{sec:intro}

Subject-driven image personalization---generating new images that faithfully preserve the identity of one or several given reference subjects in novel scenes---has become a foundational capability for modern visual content creation~\cite{ruiz2023dreambooth, gal2023textual, kumari2023multi}.
It underlies workflows ranging from on-demand product imagery and advertising creative to personalized media with consistent characters across multiple scenes to brand and IP asset generation that must remain identity-coherent across diverse contexts.
The underlying technical demand is for a single model that, given novel reference subjects and a text prompt at inference time, produces images that preserve each subject's identity across arbitrary contexts without per-subject retraining.

The dominant approach today fine-tunes a single pretrained multimodal diffusion transformer (MMDiT)~\cite{esser2024scaling, flux2} on large paired datasets of \emph{(reference, composed-target)} examples~\cite{xiao2025omnigen, mou2025dreamo, wang2025scone, wu2025omnigen2, cheng2025umo}, where each composed target is a synthesized image of the subject placed in a novel scene.
This paradigm has driven recent progress in generalized, encoder-free personalization.
However, producing the training targets demands a multi-stage curation pipeline---LLM-based prompt generation, T2I-based composed-target synthesis, reference-subject extraction, VLM-based quality filtering, and correspondence labeling~\cite{she2025mosaic, chen2025xverse, wu2025uno}---yielding $150$K to over $2$M paired samples per system and tightly coupling each method to a particular target synthesizer and curation choice.

To address this, we propose \emph{CRAFT} (\textbf{C}onstrained \textbf{R}eward via \textbf{A}ttention \textbf{F}ine-\textbf{T}uning).
We start from the observation that reference-aware MMDiTs, which jointly attend to text, noise, and reference-image tokens, often produce subject-aligned attention patterns even before any fine-tuning. This opens up an alternative to composed-target supervision: rather than inducing such routing through paired data, we shape this existing routing directly with a lightweight reward signal, guided by a \emph{Where to look} principle---at a small subset of (step, block) coordinates, noise- and prompt-token attention should attend to the correct reference region. The supervision this requires is correspondingly minimal---just reference images paired with subject masks, eliminating the need for expensive composed-target pairs.

Concretely, CRAFT operates within a single-step Reward Feedback Learning (ReFL) loop~\cite{xu2024imagereward, clark2024directly} with reward signals backpropagated into LoRA adapters~\cite{hu2022lora}. Attention-level rewards align noise- and prompt-token attention with the correct reference subject at the (step, block) coordinates identified on the unmodified backbone, realizing the \emph{Where to look} principle. The resulting per-subject attention masks then gate a pixel-level identity reward, keeping image-space supervision consistent with the learned attention routing. Trained on only $10$K reference images and subject masks, CRAFT achieves state-of-the-art performance on various benchmarks, including XVerseBench, DreamBench, and OmniContext, and the same recipe transfers to other reference-aware backbones. In summary, our contributions are as follows:

\begin{itemize}
    \item We reformulate generalized subject-driven personalization as a reference-side reward problem and introduce \textit{CRAFT}, a single-step ReFL framework that adapts a pre-trained reference-aware MMDiT via LoRA, trained solely on reference-side supervision.

    \item We introduce \emph{attention-level rewards} that realize the \emph{Where to look} principle, aligning noise- and prompt-token attention with the correct reference region at a small subset of (step, block) coordinates identified on the unmodified backbone.

    \item We tie image-space supervision to learned attention via \emph{attention-derived per-subject masks that gate a pixel-level identity reward}, keeping pixel-level supervision consistent with the routing the attention rewards produce.

    \item \textit{CRAFT} achieves state-of-the-art performance on various benchmarks using only $10$K reference-only instances \rev{and no composed-target supervision, in contrast to the $150$K to over $2$M composed-target pairs prior generalized methods require}, and the recipe transfers across reference-aware MMDiT backbones.
\end{itemize}

\section{Related Work}
\label{sec:related}

\subsection{Subject-Driven Image Personalization}

Subject-driven image personalization aims to generate novel scenes that preserve the identity of one or more reference subjects under arbitrary text prompts, building on pre-trained text-to-image diffusion models. Existing approaches can be broadly categorized into per-subject optimization, encoder-based adapters, and generalized multi-subject models. Per-subject optimization methods~\cite{ruiz2023dreambooth, gal2023textual, kumari2023multi} fine-tune model parameters for each subject, achieving high fidelity but requiring a separate optimization process per instance. Encoder-based adapters~\cite{ye2023ipadapter, li2024blip, wang2024instantid} remove this cost by injecting reference features at inference time, but are typically limited to specific domains and struggle with multi-subject composition.

Generalized methods~\cite{xiao2025omnigen, wu2025uno, mou2025dreamo, chen2025xverse, she2025mosaic, wang2025scone, wu2025omnigen2, cheng2025umo} address these limitations by training a pretrained single backbone on large paired datasets of (reference, composed-target) samples. In these approaches, composed-target images implicitly supervise how subjects should be placed and rendered, and are sometimes further used to derive explicit alignment signals~\cite{she2025mosaic}. However, this reliance on composed-target supervision requires large-scale data construction and tightly couples training to the quality of synthesized targets. In contrast, we avoid composed-target data entirely and instead directly constrain how the model attends to reference subjects during generation.

\subsection{Reward Fine-Tuning of Diffusion Models}
Reward-based fine-tuning of diffusion models has been explored to align generation with external evaluators. Policy-gradient approaches~\cite{black2023training, fan2024dpok} treat denoising as a sequential decision process, while reward-backpropagation methods~\cite{xu2024imagereward, clark2024directly, prabhudesai2023aligning} differentiate through the denoising process directly. ReFL~\cite{xu2024imagereward} and DRaFT-1~\cite{clark2024directly} backpropagate through a single denoising step for efficiency, whereas AlignProp~\cite{prabhudesai2023aligning} extends this to full trajectories at a higher computational cost.

Most prior work places rewards on the output image---for example, ImageReward~\cite{xu2024imagereward} and human-preference scores~\cite{wu2023human}---leaving the model's internal attention behavior unsupervised. In personalization, UMO~\cite{cheng2025umo} extends ReFL with a Hungarian-matched multi-identity reward, but couples it with a diffusion loss against composed target images, requiring paired data. In contrast, we place reward signals directly on the cross-modal attention sub-blocks, eliminating the need for composed-target supervision.

\section{Preliminaries}
\label{sec:pre}

\subsection{Multimodal Diffusion Transformers}
Recent text-to-image diffusion models have transitioned from U-Net-based architectures~\cite{rombach2022high} to transformer-based designs that process all modalities within a unified attention framework~\cite{esser2024scaling, flux2}. We build on FLUX.2-klein~\cite{flux2}, a $9$B-parameter multimodal diffusion transformer (MMDiT) distilled to four denoising steps. The model jointly attends to text prompt, noise, and reference image tokens within a single attention sequence. Given prompt tokens $\mathbf{P}$, noise tokens $\mathbf{N}$, and per-reference tokens $\mathbf{R}_k$ for $k \in \{1, \dots, K\}$, the model processes the concatenated sequence as follows:
\begin{equation}
    \mathbf{S} = [\mathbf{P}; \mathbf{N}; \mathbf{R}_1; \dots; \mathbf{R}_K] \in \mathbb{R}^{(N_p + N_z + K \cdot N_r) \times d},
\end{equation}
where $\mathbf{P} \in \mathbb{R}^{N_p \times d}$ are prompt tokens, $\mathbf{N} \in \mathbb{R}^{N_z \times d}$ are noise tokens, and $\mathbf{R}_k \in \mathbb{R}^{N_r \times d}$ are the tokens corresponding to the $k$-th reference image, with $d$ the shared token embedding dimension.

\noindent\textbf{Attention sub-block notation.}
At denoising step $t$ and transformer block $b$, self-attention over the concatenated sequence $\mathbf{S}=[\mathbf{P};\mathbf{N};\mathbf{R}_1;\dots;\mathbf{R}_K]$ produces
\begin{equation}
    \mathbf{A}^{t,b}
    =
    \mathrm{softmax}
    \bigl(\mathbf{Q}^{t,b}{\mathbf{K}^{t,b}}^\top/\sqrt{d}\bigr).
\end{equation}
For token groups $X,Y\in\{P,N,R_1,\dots,R_K\}$, we denote by $\mathbf{A}^{t,b}_{X2Y}$ the sub-block of $\mathbf{A}^{t,b}$ whose queries come from group $X$ and whose keys come from group $Y$. The first index denotes the query side, and the second index denotes the key side. Throughout, group labels appearing in attention subscripts (e.g., $N$, ${P}$, ${R}_k$) are kept in italic, whereas the corresponding token matrices ($\mathbf{N}$, $\mathbf{P}$, $\mathbf{R}_k$) are written in bold.

\subsection{Reward Feedback Learning (ReFL)}

Reward Feedback Learning (ReFL)~\cite{xu2024imagereward} fine-tunes diffusion models using differentiable reward signals through truncated backpropagation. In general, a reward step $t$ is sampled from a range $t \in [T_1, T_2]$, and the model backpropagates the reward through a single denoising step at $t$. The denoising trajectory from $T$ to $t{+}1$ is executed without gradients, and the reward is computed from the predicted clean image at step $t$. For flow-matching models~\cite{lipman2023flow, liu2023flow}, given the noisy latent $\mathbf{z}_t$ at step $t$, the clean image is estimated as:
\begin{equation}
    \hat{\mathbf{x}}_0 = \mathbf{z}_t - t \cdot f_\theta(\mathbf{z}_t, t),
\end{equation}
where $f_\theta$ is the predicted velocity. Single-step variants such as ReFL and DRaFT-1~\cite{clark2024directly} are well-suited for large diffusion transformers due to their favorable memory and stability properties.

\begin{figure*}[t]
    \centering
    \includegraphics[width=.92\linewidth]{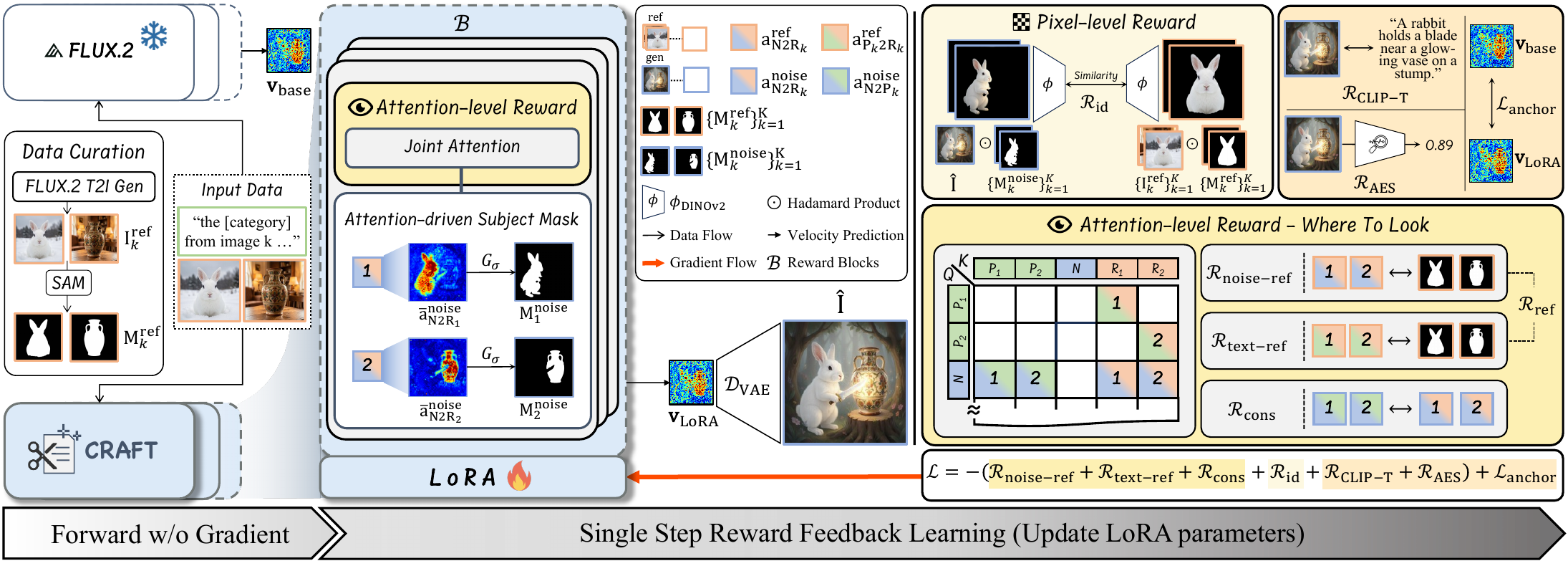}
    \caption{\textbf{Overall pipeline of CRAFT.} The denoising prefix is rolled out without gradients up to the reward step $t^\ast$, where two forward passes are performed: the frozen backbone $f_\text{base}$ produces $\mathbf{v}_\text{base}$, while the LoRA-adapted model $f_\text{LoRA}$ produces $\mathbf{v}_\text{LoRA}$ together with the cross-modal attention sub-blocks $\mathbf{A}_{N2R_k}$, $\mathbf{A}_{N2P_k}$, and $\mathbf{A}_{P_k 2 R_k}$. Following the \textit{Where to look} principle, $\mathcal{R}_\text{ref}$ and $\mathcal{R}_\text{cons}$ shape noise- and phrase-token attention toward each reference subject (\secref{sec:method:attn_rewards}); the resulting per-subject mask $\mathbf{m}^\text{noise}_k$ then gates the pixel-level identity reward $\mathcal{R}_\text{id}$ on the VAE-decoded pre-image $\hat{\mathbf{I}}$ (\secref{sec:method:pixel}). Auxiliary terms $\mathcal{R}_\text{CLIP-T}$, $\mathcal{R}_\text{AES}$, and a velocity-space regularizer $\mathcal{L}_\text{anchor}$ are added for prompt fidelity, aesthetics, and stability (\secref{sec:method:loss}). A single backward pass updates only the LoRA parameters.}
    \label{fig:main}
\end{figure*}

\section{Method}
\label{sec:method}

CRAFT fine-tunes LoRA adapters of a frozen reference-aware MMDiT using single-step Reward Feedback Learning (ReFL). Instead of supervising composed target images, CRAFT supervises the model's cross-modal routing: attention-level rewards encourage each subject's noise and phrase tokens to focus on the corresponding reference subject, and the resulting attention localization gates a pixel-level identity reward. The overall procedure is summarized in \figref{fig:main} and \algref{alg:craft}.

\subsection{Reference-Conditioned Reward Setup}
\label{sec:method:setup}

Each training instance is written as $\bigl(y,\{\mathbf{I}^\text{ref}_k, \mathbf{M}^\text{ref}_k\}_{k=1}^{K}\bigr)$, where $y$ is a prompt mentioning $K$ reference subjects, $\mathbf{I}^\text{ref}_k$ is the $k$-th reference image, and $\mathbf{M}^\text{ref}_k$ is its image-space subject mask. The masks are reward-side guidance only: the model receives $y$ and $\{\mathbf{I}^\text{ref}_k\}_{k=1}^{K}$, but never $\{\mathbf{M}^\text{ref}_k\}_{k=1}^{K}$. Thus, inference remains mask-free.
Dataset construction details are provided in \secref{app:dataset}. \rev{Briefly, references are isolated single-subject images rendered with FLUX.2~\cite{flux2} in text-to-image mode, and each mask $\mathbf{M}^\text{ref}_k$ is obtained with Grounded-SAM~\cite{ren2024grounded}; no scene composition or composed-target synthesis is involved.}
Tokenizing $y$ gives a prompt-token sequence $\mathbf{P}$. For each subject $k$, we denote by $\mathbf{P}_k\subset \mathbf{P}$ the contiguous token span of its referring phrase (e.g. ``the [category] from image k''), and by $\mathbf{R}_k$ the reference-token block extracted from $\mathbf{I}^\text{ref}_k$. For reward computation, we denote by $\mathbf{m}^\text{ref}_k\in\{0,1\}^{N_r}$ the downsampled reference-token-grid mask of $\mathbf{M}^\text{ref}_k$.

CRAFT applies attention-level rewards at a selected reward locus, specified by a denoising step $t^\ast$ and a set of transformer blocks $\mathcal{B}$. Using the attention notation from \secref{sec:pre}, we write throughout the method \rev{as follows}:
\begin{equation}
    \mathbf{A}_{X2Y}
    :=
    \frac{1}{|\mathcal{B}|}
    \sum_{b\in\mathcal{B}}
    \mathbf{A}^{t^\ast,b}_{X2Y}, \quad X, Y \in \{N, P_k, R_k\}\rev{.}
\end{equation}
The concrete choice of $t^\ast$ and $\mathcal{B}$ is determined by profiling the backbone: 
We select the (step $t^\ast$, block $\mathcal{B}$) which $\mathbf{A}_{N2R_k}$ best aligns with the generated subject, as reported in \secref{app:routing}.

CRAFT uses three subject-specific attention handles: $\mathbf{A}_{N2R_k}$, $\mathbf{A}_{N2P_k}$, $\mathbf{A}_{P_k 2 R_k}$.
Here, $\mathbf{A}_{N2R_k}$ is the attention from noise tokens to the $k$-th reference, $\mathbf{A}_{N2P_k}$ from noise tokens to $\mathbf{P}_k$, and $\mathbf{A}_{P_k 2 R_k}$ from $\mathbf{P}_k$ to the $k$-th reference.
These handles implement the \emph{Where to look} principle: the generated subject should retrieve visual evidence from the correct reference region, the corresponding phrase should ground to the same region, and the two induced noise-grid localizations should agree spatially.

\subsection{Attention-Level Rewards: \textit{Where to Look}}
\label{sec:method:attn_rewards}
Using the subject-specific attention handles defined in \secref{sec:method:setup}, we instantiate the \emph{Where to look} principle through three complementary attention-level rewards.
$\mathcal{R}_\text{noise-ref}$ aligns \emph{noise} queries with the subject region of the reference, $\mathcal{R}_\text{text-ref}$ aligns \emph{phrase} queries with the same region, and $\mathcal{R}_\text{cons}$ enforces spatial agreement between the two noise-grid localizations of subject $k$---one obtained from $\mathbf{A}_{N2R_k}$, one from $\mathbf{A}_{N2P_k}$.  Below, $\text{norm}(\cdot)$ denotes per-vector min--max normalization to $[0,1]$. Marginalized attention vectors carry a superscript indicating the surviving grid ($\text{ref}$ for $\mathbb{R}^{N_r}$, $\text{noise}$ for $\mathbb{R}^{N_z}$) and a subscript identifying the source sub-block.

\noindent\textbf{Noise--reference alignment.}
$\mathcal{R}_\text{noise-ref}$ measures, for each subject $k$, the fraction of normalized noise-to-reference response concentrated on the subject region:
\begin{equation}
\begin{gathered}
    \mathcal{R}_\text{noise-ref} = \frac{1}{K}\sum_{k=1}^{K} \frac{\sum_r \mathbf{a}^\text{ref}_{N2R_k} \odot \mathbf{m}^\text{ref}_k}{\sum_r \mathbf{a}^\text{ref}_{N2R_k} + \varepsilon},\\
    \mathbf{a}^\text{ref}_{N2R_k} = \text{norm}\!\Big(\textstyle\sum_n \mathbf{A}_{N2R_k}\Big) \;\in\; \mathbb{R}^{N_r},
    \label{eq:r_noise_ref}
\end{gathered}
\end{equation}
where $\odot$ denotes the Hadamard product, $\varepsilon$ is a small constant that avoids division by zero, and $\mathbf{m}^\text{ref}_k \in \{0,1\}^{N_r}$ is the bilinear downsampling of $\mathbf{M}^\text{ref}_k$ to the reference-token grid. Maximizing $\mathcal{R}_\text{noise-ref}$ encourages noise tokens to attend to the subject region of each reference, so the generated image tends to draw subject features from the reference subject rather than its background.

\noindent\textbf{Text--reference alignment.}
To ground subject-$k$'s phrase tokens in their visual evidence, $\mathcal{R}_\text{text-ref}$ applies the same fraction-on-mask form to the phrase side, acting on the phrase-to-reference attention $\mathbf{A}_{P_k 2 R_k}$ and marginalizing over the phrase-query axis:
\begin{equation}
\begin{gathered}
    \mathcal{R}_\text{text-ref} = \frac{1}{K}\sum_{k=1}^{K} \frac{\sum_r \mathbf{a}^\text{ref}_{P_k 2 R_k} \odot \mathbf{m}^\text{ref}_k}{\sum_r \mathbf{a}^\text{ref}_{P_k 2 R_k} + \varepsilon},\\
    \mathbf{a}^\text{ref}_{P_k 2 R_k} = \text{norm}\!\Big(\textstyle\sum_{p} \mathbf{A}_{P_k 2 R_k}\Big) \;\in\; \mathbb{R}^{N_r}.
    \label{eq:r_text_ref}
\end{gathered}
\end{equation}
$\mathcal{R}_\text{text-ref}$ aligns the textual semantics of $\mathbf{P}_k$ with the visual content inside $\mathbf{R}_k$, grounding each phrase token in the reference subject rather than the generic semantics carried by the text encoder.

Since $\mathcal{R}_\text{noise-ref}$ and $\mathcal{R}_\text{text-ref}$ both pull queries toward the same reference subject region of $\mathbf{m}^\text{ref}_k$, we bundle them into a single reference-mask alignment term used by the training objective (\secref{sec:method:loss}):
\begin{equation}
    \mathcal{R}_\text{ref} = w_\text{nr}\,\mathcal{R}_\text{noise-ref} + w_\text{tr}\,\mathcal{R}_\text{text-ref},
    \label{eq:r_ref}
\end{equation}
where $w_\text{nr}, w_\text{tr} \geq 0$ are the weights of the noise-reference and text-reference alignment rewards, respectively.

\noindent\textbf{Text--noise spatial consistency.}
$\mathcal{R}_\text{cons}$ enforces consistent localization of subject $k$ on the \emph{noise grid} where it is generated. Noise-to-text attention is widely used as a spatial-control handle in T2I editing and guidance~\cite{hertz2022prompt, chefer2023attend, rassin2024linguistic, xiao2024fastcomposer}; in our setting, $\mathbf{P}_k$ and $\mathbf{R}_k$ both describe the same subject (through text and reference, respectively), each producing its own noise-grid localization via $\mathbf{A}_{N2P_k}$ and $\mathbf{A}_{N2R_k}$. When these two localizations land at different positions, subject $k$'s attention can split across separate regions. To prevent this, we project both blocks onto the noise grid---$\mathbf{A}_{N2R_k}$ via mask-weighted reduction with $\mathbf{m}^\text{ref}_k$, and $\mathbf{A}_{N2P_k}$ via marginalization over the phrase-key axis:
\begin{equation}
\begin{gathered}
    \mathbf{a}^\text{noise}_{N2R_k} = \text{norm}\!\Big(\textstyle\sum_r \mathbf{A}_{N2R_k} \odot \mathbf{m}^\text{ref}_k\Big) \;\in\; \mathbb{R}^{N_z},\\
    \mathbf{a}^\text{noise}_{N2P_k} = \text{norm}\!\Big(\textstyle\sum_{p} \mathbf{A}_{N2P_k}\Big) \;\in\; \mathbb{R}^{N_z},
    \label{eq:h_defs}
\end{gathered}
\end{equation}
and define $\mathcal{R}_\text{cons}$ as their probabilistic soft IoU:
\begin{equation}
    \mathcal{R}_\text{cons} = \frac{1}{K}\sum_{k=1}^{K} \frac{\sum_n \mathbf{a}^\text{noise}_{N2R_k}  \mathbf{a}^\text{noise}_{N2P_k}}{\sum_n \big(\mathbf{a}^\text{noise}_{N2R_k} + \mathbf{a}^\text{noise}_{N2P_k} - \mathbf{a}^\text{noise}_{N2R_k}  \mathbf{a}^\text{noise}_{N2P_k}\big) + \varepsilon},
    \label{eq:r_cons}
\end{equation}
which is maximized only when the two views coincide. $\mathcal{R}_\text{cons}$ thus encourages spatial agreement between them, reducing the duplicated or misplaced subjects that can otherwise arise.

\subsection{Attention-Gated Pixel-Level Identity Reward}
\label{sec:method:pixel}
\label{sec:mask}

The attention-level supervision in \secref{sec:method:attn_rewards} shapes routing inside the transformer. We complement it with a pixel-space identity reward $\mathcal{R}_\text{id}$ on the decoded pre-image $\hat{\mathbf{I}} = \mathcal{D}_\text{VAE}(\mathbf{z}_{t^\ast} - t^\ast\, \mathbf{v}_\text{LoRA})$, where $\mathcal{D}_\text{VAE}$ is the FLUX.2 VAE decoder. To keep this pixel-level signal aligned with the routing being optimized, we measure identity only where the attention has placed the subject by gating $\mathcal{R}_\text{id}$ with a per-subject mask derived from the same attention.

\noindent\textbf{Attention-derived subject mask.}
Since $\mathbf{A}_{N2R_k}$ encodes where subject $k$ lands on the noise grid, we reuse it directly as a localization signal. As supported by the subject-routing analysis in \secref{app:routing}, $\mathbf{A}_{N2R_k}$ already aligns reasonably well with the generated subject region in the unmodified backbone, and $\mathcal{R}_\text{noise-ref}$ and $\mathcal{R}_\text{cons}$ further sharpen this alignment during training, making external re-segmentation of the generated image unnecessary. Gaussian-smoothing the mask-weighted sum and thresholding at $0.5$ yields the per-subject mask:
\begin{equation}
\begin{gathered}
 \mathbf{m}^\text{noise}_k =\begin{cases} 1 & \text{if } \bar{\mathbf{a}}^\text{noise}_{N2R_k} > 0.5 \\ 0 & \text{otherwise} \end{cases} \in \{0,1\}^{N_z},\\
    \bar{\mathbf{a}}^\text{noise}_{N2R_k} = \text{norm}\!\Big(G_\sigma\!\Big(\textstyle\sum_r \mathbf{A}_{N2R_k} \odot \mathbf{m}^\text{ref}_k\Big)\Big) \;\in\; \mathbb{R}^{N_z},
    \label{eq:m_attn}
\end{gathered}
\end{equation}
where $G_\sigma$ is a 2D Gaussian smoothing kernel with $\sigma{=}2$, and the case definition is applied element-wise.

\noindent\textbf{Identity reward.}
$\mathcal{R}_\text{id}$ scores DINOv2~\cite{oquab2024dinov2} similarity between the generated subject and its reference, gated by $\mathbf{m}^\text{noise}_k$ so that identity is measured exactly where the attention has placed the subject in $\hat{\mathbf{I}}$:
\begin{equation}
\mathcal{R}_\text{id} = \frac{1}{K}\sum_{k=1}^{K}
    \cos\!\big(\phi_\text{DINOv2}(\hat{\mathbf{I}}\odot \mathbf{M}^\text{noise}_k),\;
               \phi_\text{DINOv2}(\mathbf{I}^\text{ref}_k\odot \mathbf{M}^\text{ref}_k)\big),
    \label{eq:r_id}
\end{equation}
where $\phi_\text{DINOv2}$ is a frozen image encoder and $\mathbf{M}^\text{noise}_k$ is the bilinear upsampling of $\mathbf{m}^\text{noise}_k$ to the image grid.

Due to $\mathbf{m}^\text{noise}_k$ is derived from $\mathbf{A}_{N2R_k}$---which $\mathcal{R}_\text{noise-ref}$ shapes directly and $\mathcal{R}_\text{cons}$ couples to $\mathbf{A}_{N2P_k}$---the gate moves with the attention being optimized. As the attention rewards sharpen this gate, $\mathcal{R}_\text{id}$ scores identity over a tighter region around the actual subject, so the attention and pixel supervisions consistently target the same spatial location rather than disagreeing on where the subject should appear.

\begin{algorithm}[t]
\caption{CRAFT training (single-step ReFL).}
\label{alg:craft}
\begin{algorithmic}[1]
\REQUIRE Base model $f_\text{base}$ (frozen) and LoRA-adapted model $f_\text{LoRA}$ with learnable parameters $\Delta\theta$, reward step $t^\ast$, reward blocks $\mathcal{B}$, learning rate $\eta$.
\FOR{each training instance $(y, \{\mathbf{I}^\text{ref}_k, \mathbf{M}^\text{ref}_k\}_{k=1}^{K})$}
    \STATE Sample $\mathbf{z}_T \sim \mathcal{N}(\mathbf{0}, \mathbf{I})$.
    \FOR{$\tau = T, T{-}1, \dots, t^\ast{+}1$}
        \STATE \textbf{no grad:} $\mathbf{z}_{\tau-1} \gets \mathbf{z}_\tau - \tau\, f_\text{base}(\mathbf{z}_\tau, \tau \mid y, \{\mathbf{I}^\text{ref}_k\})$ \hfill \textcolor{blue}{// roll out denoising prefix}
    \ENDFOR
    \STATE \textbf{no grad:} $\mathbf{v}_\text{base} \gets f_\text{base}(\mathbf{z}_{t^\ast}, t^\ast \mid y, \{\mathbf{I}^\text{ref}_k\})$ \hfill \textcolor{blue}{// base velocity}
    \STATE \textbf{with grad:} $\mathbf{v}_\text{LoRA},\, \mathcal{A} \gets f_\text{LoRA}(\mathbf{z}_{t^\ast}, t^\ast \mid y, \{\mathbf{I}^\text{ref}_k\})$ \hfill \textcolor{blue}{// $\mathcal{A} = \{\mathbf{A}_{N2R_k}, \mathbf{A}_{N2P_k}, \mathbf{A}_{P_k 2 R_k}\}_{k=1}^{K}$}
    \STATE Decode the pre-image: $\hat{\mathbf{I}} \gets \mathcal{D}_\text{VAE}(\mathbf{z}_{t^\ast} - t^\ast\, \mathbf{v}_\text{LoRA})$.
    \STATE Compute attention rewards $\mathcal{R}_\text{ref}, \mathcal{R}_\text{cons}$ and per-subject masks $\{\mathbf{m}^\text{noise}_k\}$ via Equations~\eqref{eq:r_noise_ref}-\eqref{eq:m_attn}.
    \STATE Compute identity reward $\mathcal{R}_\text{id}$ via \equref{eq:r_id}; auxiliary terms $\mathcal{R}_\text{CLIP-T}, \mathcal{R}_\text{AES}, \mathcal{L}_\text{anchor}$ as defined in \secref{sec:method:loss}.
    \STATE Form the total loss $\mathcal{L}$ via \equref{eq:total_loss} and update $\Delta\theta \gets \Delta\theta - \eta\, \nabla_{\Delta\theta}\, \mathcal{L}$.
\ENDFOR
\end{algorithmic}
\end{algorithm}

\subsection{Training Objective}
\label{sec:method:loss}
We combine the bundled reference-mask alignment $\mathcal{R}_\text{ref}$ (\equref{eq:r_ref}), the spatial consistency $\mathcal{R}_\text{cons}$ (\equref{eq:r_cons}), and the pixel identity reward $\mathcal{R}_\text{id}$ (\equref{eq:r_id}) into the CRAFT objective. We further include two auxiliary reward terms---a CLIP~\cite{radford2021learning} text--image similarity $\mathcal{R}_\text{CLIP-T}$ and an aesthetic predictor~\cite{improved_aesthetic} score $\mathcal{R}_\text{AES}$---for prompt fidelity and aesthetic quality, together with a velocity-space anchor $\mathcal{L}_\text{anchor} = \|\mathbf{v}_\text{LoRA} - \text{sg}(\mathbf{v}_\text{base})\|_2^2$ that ties the LoRA velocity field to the base model for training stability. The full training objective is as follows:
\begin{equation}
\begin{aligned}
    \mathcal{L} = & -\mathcal{R}_\text{ref} - w_\text{c}\,\mathcal{R}_\text{cons} - w_\text{id}\,\mathcal{R}_\text{id} \\
                  & - w_\text{t}\,\mathcal{R}_\text{CLIP-T} - w_\text{a}\,\mathcal{R}_\text{AES} + w_\text{anchor}\,\mathcal{L}_\text{anchor},
\end{aligned}
    \label{eq:total_loss}
\end{equation}
where the reward terms enter with negative signs so that gradient descent maximizes them. \rev{The first three terms---$\mathcal{R}_\text{ref}$, $\mathcal{R}_\text{cons}$, and $\mathcal{R}_\text{id}$---are the essential CRAFT-specific rewards, and are the ones ablated in \tabref{tab:ablation}; the remaining three ($\mathcal{R}_\text{CLIP-T}$, $\mathcal{R}_\text{AES}$, $\mathcal{L}_\text{anchor}$) are auxiliary stabilizers inherited from standard reward fine-tuning.} Weights $w_\text{c}, w_\text{id}, w_\text{t}, w_\text{a}, w_\text{anchor} \geq 0$. \algref{alg:craft} summarizes the full procedure.

\section{Experiments}
\label{sec:experiment}

\begin{table*}[t]
\centering\small
\caption{
Quantitative comparison on XVerseBench. ``Target.'' marks composed-target supervision (\Checkmark/\XSolidBrush); ``\#Data'' is the reported training-sample count. $^{\dag}$ denotes results reproduced with official code; (\textit{---}) denotes not reported. Best/second-best in \textbf{bold}/\underline{underlined}. $^{\ast}$ \emph{CRAFT (mask-free)} operates on raw, unsegmented references at inference and is reported for reference (not included in the ranking). \rev{Because masks are reward-side training annotations only, this mask-free setting is CRAFT's intended inference mode, and it performs on par with or above the segmented-input configuration (Overall $77.80$ vs.\ $76.47$).}
}
\label{tab:xversebench}
\resizebox{\linewidth}{!}{
\begin{tabular}{lcc|ccccc|ccccc|c}
\toprule
& & & \multicolumn{5}{c|}{\textbf{Single subject (90 prompts)}}
& \multicolumn{5}{c|}{\textbf{Multi subject (210 prompts)}}
& \multirow{2}{*}{\textbf{Overall}~$\uparrow$} \\
Method & Target. & \#Data & DPG~$\uparrow$ & ID~$\uparrow$ & IP~$\uparrow$ & AES~$\uparrow$ & AVG~$\uparrow$
       & DPG~$\uparrow$ & ID~$\uparrow$ & IP~$\uparrow$ & AES~$\uparrow$ & AVG~$\uparrow$
       & \\
\midrule
UNO~\cite{wu2025uno}                       & \Checkmark & 245K  & 89.65 & 47.91 & 80.40 & 55.90 & 68.47 & 85.28 & 31.82 & 67.00 & 54.24 & 59.59 & 64.03 \\
OmniGen~\cite{xiao2025omnigen}             & \Checkmark & 6.5M  & 83.90 & 76.51 & 78.46 & 51.41 & 72.57 & 78.23 & 55.53 & 62.32 & 49.84 & 61.48 & 67.03 \\
OmniGen2~\cite{wu2025omnigen2}             & \Checkmark & 180K  & 92.60 & 62.41 & 74.08 & 52.34 & 70.36 & \textbf{91.55} & 40.81 & 67.15 & 51.40 & 62.73 & 66.55 \\
DreamO~\cite{mou2025dreamo}                & \Checkmark & 150K  & \textbf{96.93} & 75.48 & 70.84 & 54.57 & 74.46 & 88.80 & 50.24 & 64.63 & 52.47 & 64.04 & 69.25 \\
UMO$^{\dag}$~\cite{cheng2025umo}                & \Checkmark & \textit{---}  & 86.75 & 77.36 & 76.99 & \textbf{61.71} & 75.70 & 87.18 & 58.24 & 60.52 & \textbf{58.74} & 66.17 & 70.94 \\
XVerse~\cite{chen2025xverse}               & \Checkmark & $2$M$+$ & 93.69 & 79.48 & 76.86 & 56.84 & 76.72 & 88.26 & \underline{66.59} & 71.48 & 53.97 & 70.08 & 73.40 \\
MOSAIC~\cite{she2025mosaic}                & \Checkmark & 1.2M  & 96.55 & \underline{81.98} & \underline{80.92} & 60.77 & \underline{80.05} & \underline{88.94} & \textbf{69.90} & \underline{74.27} & 55.02 & \textbf{72.03} & \underline{76.04} \\
\midrule
\textbf{CRAFT (Ours)}                      & \XSolidBrush & \textbf{10K}   & \underline{96.81} & \textbf{84.22} & \textbf{84.23} & \underline{61.24} & \textbf{81.62} & 88.71 & 61.16 & \textbf{77.25} & \underline{58.16} & \underline{71.32} & \textbf{76.47} \\
\textit{CRAFT (mask-free, Ours)}$^{\ast}$  & \XSolidBrush & \textbf{10K}   & \textit{98.91} & \textit{87.67} & \textit{86.74} & \textit{62.58} & \textit{83.97} & \textit{89.12} & \textit{60.35} & \textit{77.28} & \textit{59.71} & \textit{71.62} & \textit{77.80} \\
\bottomrule
\end{tabular}%
}
\end{table*}

\begin{figure*}[t]
    \centering
    \includegraphics[width=.9\linewidth]{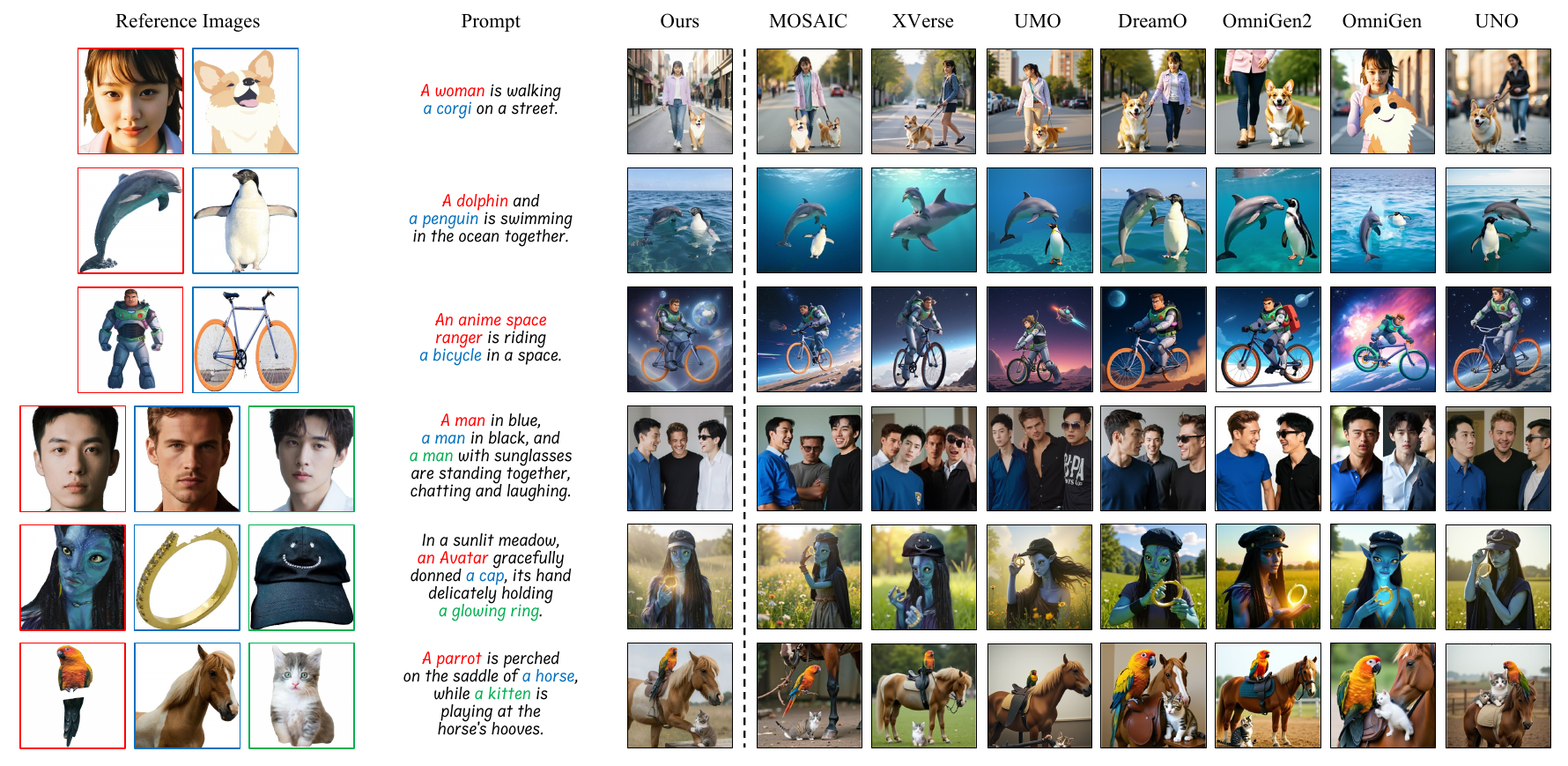}
    \caption{Qualitative comparison with state-of-the-art subject-driven personalization models.}
    \label{fig:comparison}
\end{figure*}

\subsection{Implementation Details}
\label{sec:exp_impl}

\noindent \textbf{Main backbone and training.}
We implement CRAFT on FLUX.2-klein~\cite{flux2}, a $9$B-parameter MMDiT distilled to four denoising steps, with the backbone frozen and learning concentrated in LoRA adapters. We attach LoRA modules (rank $r = 64$) to all attention layers and train them at resolution $1024^2$ on four NVIDIA B200 GPUs using AdamW with learning rate $2{\times}10^{-6}$ (constant schedule with $50$ warmup steps). 

\noindent \textbf{Reward locus.}
Computing attention-level rewards requires reading attention tensors, which disables Flash Attention~\cite{dao2022flashattention, dao2023flashattention2}; applying such supervision to all steps and blocks is therefore inefficient. We select a compact reward locus by running a subject-routing analysis on the unmodified FLUX.2-klein backbone and choosing the step/block coordinates whose noise-to-reference attention best aligns with generated subject masks. The full protocol is described in \secref{app:routing}. For FLUX.2-klein, this analysis identifies
\begin{equation}
    t^\ast=2,\qquad
    \mathcal{B}
    =
    \{\texttt{single\_1},\texttt{single\_9},\texttt{single\_8}\}.
\end{equation}

\noindent \textbf{Loss weights and data.}
We set \(w_\text{nr}=w_\text{tr}=0.5\), \(w_\text{c}=1.0\), \(w_\text{id}=1.0\), \(w_\text{t}=0.3\), \(w_\text{a}=3\times10^{-3}\), and \(w_\text{anchor}=0.5\). Training runs for 3,000 optimizer steps. The training dataset contains 10,000 reference-only instances; its construction is detailed in \secref{app:dataset}.

\subsection{Evaluation Setup}
\label{sec:exp:eval}

\noindent\textbf{Benchmark.}
We follow the official evaluation protocol of XVerseBe\\nch~\cite{chen2025xverse}, an extension of DreamBench++~\cite{peng2025dreambenchpp} designed to comprehensively assess single- and multi-subject controllable generation. XVerseBench augments the DreamBench++ dataset with $20$ newly generated portrait images, resulting in a benchmark that comprises $20$ distinct human identities, $74$ unique objects, and $45$ animal species and individuals. The benchmark contains $300$ test prompts spanning single-, dual-, and triple-subject combinations across humans, objects, and animals; following the official split, we evaluate on $90$ single-subject and $210$ multi-subject (dual~$+$~triple) prompts. The official protocol provides segmented reference images at evaluation time, which we use for the main comparison. We also reported additional benchmark results on DreamBench~\cite{ruiz2023dreambooth} and OmniContext~\cite{wu2025omnigen2} in \secref{app:additional_benchmarks}.

\noindent\textbf{Evaluation Metrics.}
Following XVerseBench~\cite{chen2025xverse}, we report four complementary metrics: \emph{DPG} for prompt fidelity (mPLUG-Owl~\cite{ye2023mplug} answering Davidsonian Scene Graph \cite{cho2023davidsonian} questions on the output), \emph{ID} for human identity (ArcFace~\cite{deng2019arcface} similarity between generated and reference faces), \emph{IP} for object and animal identity (DINOv2~\cite{oquab2024dinov2} similarity over the Florence-2~\cite{xiao2024florence}/SAM~2~\cite{ravi2024sam} segmented subject), and \emph{AES} for image quality (Aesthetic Predictor v2.5~\cite{improved_aesthetic}). For a more comprehensive evaluation, we further report the four-metric average $\mathrm{AVG} = (\mathrm{DPG} + \mathrm{ID} + \mathrm{IP} + \mathrm{AES})/4$ used by XVerseBench, together with a cross-split aggregate $\mathrm{Overall} = (\mathrm{Single AVG} + \mathrm{Multi AVG})/2$.

\subsection{Comparison with State-of-the-Art Subject-Driven Personalization Models}
\label{sec:exp:comparison}

\noindent\textbf{Quantitative comparison.}
\tabref{tab:xversebench} compares CRAFT with recent subject-driven personalization methods on XVerseBench. CRAFT achieves the best Overall score ($76.47$) while using only $10$K reference-only training instances and no composed-target supervision. \rev{Following the official protocol, we average the results over four samples; the run-to-run spread is small ($\pm 0.08$ Overall), far below our lead over the next-best method.} On the single-subject split, CRAFT obtains the best AVG score ($81.62$), improving over the strongest prior method, MOSAIC~\cite{she2025mosaic}, by $+1.57$ points. The gains are especially pronounced in identity-related metrics: CRAFT achieves the highest Single ID ($84.22$) and Single IP ($84.23$), indicating that the learned routing improves both human and non-human subject preservation.

On multi-subject prompts, CRAFT achieves the best IP score ($77.25$) and the second-best Multi AVG ($71.32$). Its Multi ID is lower than MOSAIC's ($61.16$ vs.\ $69.90$), but CRAFT remains competitive overall because it better preserves object/animal identity and image quality. \rev{This multi-subject gap is in part a deliberate operating point---raising $w_\text{id}$ recovers Multi ID to $68.83$ ($\approx$ MOSAIC) at a small quality cost (\secref{app:reward_weights})---and we discuss it further in \secref{app:limitations}.} Compared with the frozen FLUX.2-klein backbone reported in \tabref{tab:ablation}, CRAFT improves Overall by $+5.43$ points, with the largest gains in subject identity preservation. These results suggest that CRAFT's reference-side rewards provide strong supervision even without paired composed targets. We also report CRAFT (mask-free), which operates on raw, unsegmented references at inference and achieves Overall $77.80$, demonstrating that CRAFT does not require test-time reference segmentation; full details are in \secref{app:mask_free}. Additional quantitative results on DreamBench~\cite{ruiz2023dreambooth} and OmniContext~\cite{wu2025omnigen2} are reported in \secref{app:additional_benchmarks}.

\begin{table*}[t]
\centering\small
\caption{
Component ablations on XVerseBench. (a) FLUX.2-klein~\cite{flux2} backbone; (b) $+\mathcal{R}_\text{ref}$\rev{;} (c) $+\mathcal{R}_\text{cons}$ (cumulative); (d) $\mathcal{R}_\text{id}$ alone on the backbone; (e) full CRAFT (all three rewards). \Checkmark marks an enabled reward. Best/second-best in \textbf{bold}/\underline{underlined}.
}
\label{tab:ablation}
\begin{tabular}{c|ccc|ccccc|ccccc|c}
\toprule
& \multicolumn{3}{c|}{\textbf{Component}}
& \multicolumn{5}{c|}{\textbf{Single subject}}
& \multicolumn{5}{c|}{\textbf{Multi subject}}
& \multirow{2}{*}{\textbf{Overall}~$\uparrow$} \\
& $\mathcal{R}_\text{ref}$ & $\mathcal{R}_\text{cons}$ & $\mathcal{R}_\text{id}$
& DPG~$\uparrow$ & ID~$\uparrow$ & IP~$\uparrow$ & AES~$\uparrow$ & AVG~$\uparrow$
& DPG~$\uparrow$ & ID~$\uparrow$ & IP~$\uparrow$ & AES~$\uparrow$ & AVG~$\uparrow$
& \\
\midrule
(a) &              &              &              & \textbf{97.22} & 66.29 & 78.15 & 59.76 & 75.36 & {\underline{88.91}} & 50.22 & 71.23 & 56.52 & 66.72 & 71.04 \\
(b) & \Checkmark   &              &              & \underline{96.89} & 70.15 & 79.75 & 60.84 & 76.91 & \textbf{88.94} & 48.11 & \rev{72.69} & 57.86 & 66.90 & 71.91 \\
(c) & \Checkmark   & \Checkmark   &              & \textbf{97.22} & 70.63 & 81.19 & \underline{61.94} & 77.75 & 88.09 & 50.43 & 70.70 & \underline{58.71} & \rev{66.98} & \rev{72.37} \\
(d) &    &    &  \Checkmark  & 96.67 & \underline{73.27} & \underline{81.93} & \textbf{61.96} & \underline{78.46} & \rev{87.32} & \rev{\underline{50.77}} & \rev{\underline{72.90}} & \rev{\textbf{60.24}} & \rev{\underline{67.81}} & \rev{\underline{73.14}} \\
\midrule
(e) & \Checkmark   & \Checkmark   & \Checkmark   & 96.81 & \textbf{84.22} & \textbf{84.23} & 61.24 & \textbf{81.62} & {88.71} & \textbf{61.16} & \textbf{77.25} & 58.16 & \textbf{71.32} & \textbf{76.47} \\
\bottomrule
\end{tabular}%
\end{table*}

\begin{table}[t]
\centering\small
\caption{
CRAFT applied to UNO~\cite{wu2025uno} and UMO~\cite{cheng2025umo} on XVerseBench. ``Target.'' marks composed-target supervision (\Checkmark/\XSolidBrush); $^{\dag}$ denotes results reproduced with official code. Best in \textbf{bold}.
}
\label{tab:uno_craft}
\resizebox{\linewidth}{!}{
\begin{tabular}{lc|c|c|c}
\toprule
Method & \textbf{Target.}
& {\textbf{Single AVG} $\uparrow$}
& {\textbf{Multi AVG} $\uparrow$}
& {\textbf{Overall}~$\uparrow$} \\
\midrule
UNO~\cite{wu2025uno}              & \Checkmark    & 68.47 & 59.59 & 64.03 \\
\textbf{UNO + Ours}               & \XSolidBrush  & \textbf{76.61} & \textbf{67.04} & \textbf{71.83} \\
\midrule
UMO$^{\dag}$~\cite{cheng2025umo}  & \Checkmark    & 75.70 & 66.17 & 70.94 \\
\textbf{UMO + Ours}               & \XSolidBrush  & \textbf{79.75}    & \textbf{69.49}    & \textbf{74.62}    \\
\bottomrule
\end{tabular}%
}
\end{table}

\begin{table}[t]
    \centering
    \footnotesize
    \caption{User study preference (\%).}
    \label{tab:user_study}
    \begin{tabular}{lccc}
    \toprule
    Method & IC $\uparrow$ & PF $\uparrow$ & IQ $\uparrow$ \\ \midrule
    UMO \cite{cheng2025umo} & 15.0 & 5.9 & 26.0 \\
    XVerse \cite{chen2025xverse} & 15.0 & 18.3 & 16.5\\
    MOSAIC \cite{she2025mosaic} & 14.8 & 13.2 & 23.3\\
    \midrule
    \textbf{Ours} & \textbf{55.2} & \textbf{62.6} &
    \textbf{34.2}\\ \bottomrule
    \end{tabular}
\end{table}

\noindent\textbf{Qualitative comparison.}
\figref{fig:comparison} compares CRAFT with seven baselines on six prompts. Two patterns stand out.

\textit{Natural human--subject interaction.}
Rows~1, 3, and 5 require a human or character to actively interact with another referenced subject: a woman walking a corgi (row~1), an anime space ranger riding a bicycle (row~3), and an Avatar wearing a cap while holding a glowing ring (row~5). In these cases, CRAFT places the subjects in compatible positions, orientations, and scales, so that the prompted action is visually supported rather than merely implied. The woman and dog appear as a coherent walking pair, the character is physically situated on the bicycle, and the cap and ring are integrated with the Avatar instead of being detached or treated as unrelated objects. Competing methods often preserve some reference appearance but fail at this relational composition: they drop one subject, place the subjects in disconnected regions, or distort one subject to accommodate the other, thereby breaking the prompted interaction.

\textit{Multi-subject identity vs.\ reference replication.}
Row~4, which asks for three men chatting, helps interpret the Multi ID gap between CRAFT and MOSAIC.
MOSAIC achieves a higher automatic face-ID score, but the visual comparison suggests that part of this score may come from reproducing the reference photographs themselves rather than recomposing the identities into the requested scene.
For example, the tight portrait framing of the references is carried into the generated image, leaving faces unnaturally cropped within an otherwise full-body group composition. CRAFT, despite being trained without composed-target supervision, instead adapts each identity to a shared group setting: the subjects are placed at compatible scales and viewpoints, and the result reads as a coherent conversation scene rather than a collage of reference portraits. This behavior is consistent with our objective: CRAFT encourages reference evidence to be routed to the correct generated subject, while still allowing the subject to be reposed and recomposed according to the prompt.

\subsection{Ablation Study}
\label{sec:exp:ablation}
\tabref{tab:ablation} isolates the effect of CRAFT's three reward components. The ablation reveals three main findings.
First, reference-mask alignment improves subject grounding. Adding $\mathcal{R}_\text{ref}$ to the frozen backbone increases Overall from $71.04$ to $71.91$, with the largest gain on the single-subject split.
This indicates that pulling both noise and phrase queries toward the reference subject region helps the model extract subject evidence from the correct part of the reference image.

Second, spatial consistency is most useful when multiple subjects compete for image space. Adding $\mathcal{R}_\text{cons}$ further improves Overall to $72.37$ and increases Multi ID from $48.11$ to $50.43$. This supports our design: aligning the noise-grid localizations induced by noise-to-reference and noise-to-text attention reduces split or misplaced subject routing.

Third, \rev{the attention rewards and the pixel identity reward are complementary. Using $\mathcal{R}_\text{id}$ alone (d) already improves over the backbone on both splits (Multi IP $71.23{\to}72.90$, Multi ID $50.22{\to}50.77$); adding the attention-shaping rewards $\mathcal{R}_\text{ref}$ and $\mathcal{R}_\text{cons}$ in the full model (e) sharpens the attention-derived gate and yields a further jump to $76.47$ Overall---Single ID to $84.22$, Multi ID to $61.16$, and Multi IP to $77.25$. Image-space identity supervision is thus most effective when the gate that localizes it is itself shaped by the attention rewards.}

\subsection{Generalization Across Reference-Aware MMDiT Backbones}
\label{sec:exp:other-backbone}
To verify that CRAFT is not specific to FLUX.2-klein, we apply the same recipe to UNO~\cite{wu2025uno} (a FLUX.1-dev-based reference-aware MMDiT) and additionally compose it on top of UMO~\cite{cheng2025umo}; per-metric numbers and implementation details are in \secref{app:uno}.

As shown in \tabref{tab:uno_craft}, \textbf{UNO + CRAFT---supervised entirely by reference-side rewards---outperforms UMO}, which is itself a reward-based fine-tuning of the same UNO backbone but requires composed-target supervision. As a result, the two rows without composed-target supervision in \tabref{tab:uno_craft} occupy the top two positions, ahead of both composed-target--supervised baselines. Composing CRAFT on top of UMO's identity-tuned LoRA yields a further consistent gain, making UMO + CRAFT the strongest UNO-family configuration we evaluate. Together with the FLUX.2-klein result in \tabref{tab:ablation}, this confirms that CRAFT generalizes across the reference-aware MMDiT backbones we evaluate, complementary to existing reward-based fine-tuning, all without introducing composed-target supervision. 

\subsection{User study}
\label{sec:user_study}
We conduct a user study with 119 participants evaluating \textit{Identity Consistency} (IC), \textit{Prompt Fidelity} (PF), and \textit{Image Quality} (IQ).
For each question, participants are shown image sets generated by four methods in randomized order and asked to select the best option (see \secref{app:user_study_details} for the detailed format).
We compare with UMO, XVerse, and MOSAIC, the top three quantitative performers.
As shown in \tabref{tab:user_study}, our method achieves the highest preference in Identity Consistency (55.2\%), Prompt Fidelity (62.6\%), and Image Quality (34.2\%), demonstrating its effectiveness.

\section{Conclusion}
\label{sec:conclusion}
We presented \textit{CRAFT}, a reward-based fine-tuning framework for generalized subject-driven personalization that adapts a pre-trained reference-aware MMDiT via single-step ReFL with ref-side-only supervision. CRAFT realizes a \emph{Where to look} principle through three attention-level rewards---noise--reference, phrase--reference, and noise-grid spatial consistency---computed at a compact subset of (step, block) coordinates identified on the unmodified reference-aware MMDiT backbone. The same attention rewards yield per-subject masks on the noise grid that gate a pixel-level identity reward, keeping image-space supervision consistent with the learned attention routing. Trained on only $10$K self-synthesized prompt-reference instances, CRAFT achieves state-of-the-art performance across various benchmarks \rev{without any composed-target supervision, which prior methods require at the scale of $150$K to $2$M pairs}. The same recipe further transfers across reference-aware MMDiT backbones, generalizing beyond a single backbone.

\rev{\noindent\textbf{Limitations.}
CRAFT has three main limitations. First, CRAFT's clearest gains are on single-subject prompts; on multi-subject scenes, its advantage is smaller---it still attains the best multi-subject IP, but Multi-ID trails the strongest composed-target baseline (\tabref{tab:xversebench}). This gap is in part a deliberate identity--quality operating point rather than a hard limit---raising the identity weight closes much of it at a small quality cost (\secref{app:reward_weights})---and we leave stronger multi-subject identity preservation to future work. Second, CRAFT requires a backbone that natively accepts reference tokens, so text-only T2I models need a reference-conditioning module first. Third, as a reward-based recipe, it amplifies the backbone's existing routing rather than rebuilding it, so absolute performance is bounded by the base model. We detail these, together with a robustness analysis of CRAFT's design choices, in \secref{app:limitations} and \secref{app:reward_weights}.}

\begin{acks}
This work was supported by the National Research Foundation of Korea (NRF) grant funded by the Korea government (MSIT) (RS-2026-25480865) (25\%), the 2026 Cultural Technology Research and Development Project of the Ministry of Culture, Sports and Tourism and the Korea Creative Content Agency (Project name: Development of high-precision hand-crafting and service technologies based on physical AI, number: RS-2026-25510431) (25\%), the Institute of Information \& Communications Technology Planning \& Evaluation (IITP) grant funded by the Korea government (MSIT) (No.\ RS-2025-02219277, AI Star Fellowship Support Project (DGIST)) (25\%), and the ``Advanced GPU Utilization Support Program funded by the Government of the Republic of Korea (Ministry of Science and ICT) (02-26-01-0057)'' (25\%).
\end{acks}

\clearpage

\bibliographystyle{ACM-Reference-Format}
\bibliography{sample-bibliography}
\begin{figure*}
    \centering
    \includegraphics[width=0.78\linewidth]{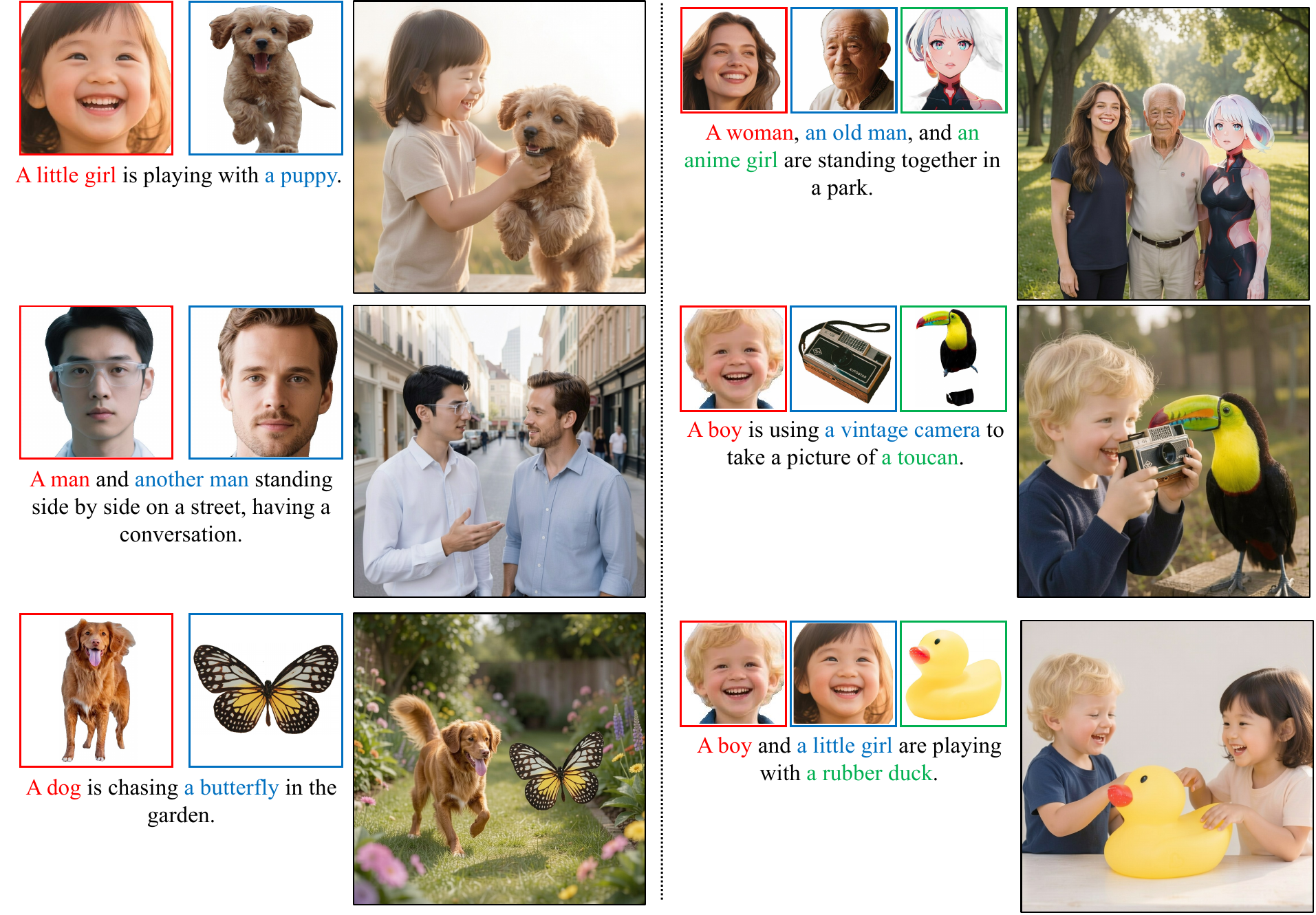}
    \caption{Qualitative results on XVerseBench under the official segmented-reference protocol.}
    \label{fig:xverse_seg}
\end{figure*}

\begin{figure*}
    \centering
    \includegraphics[width=0.78\linewidth]{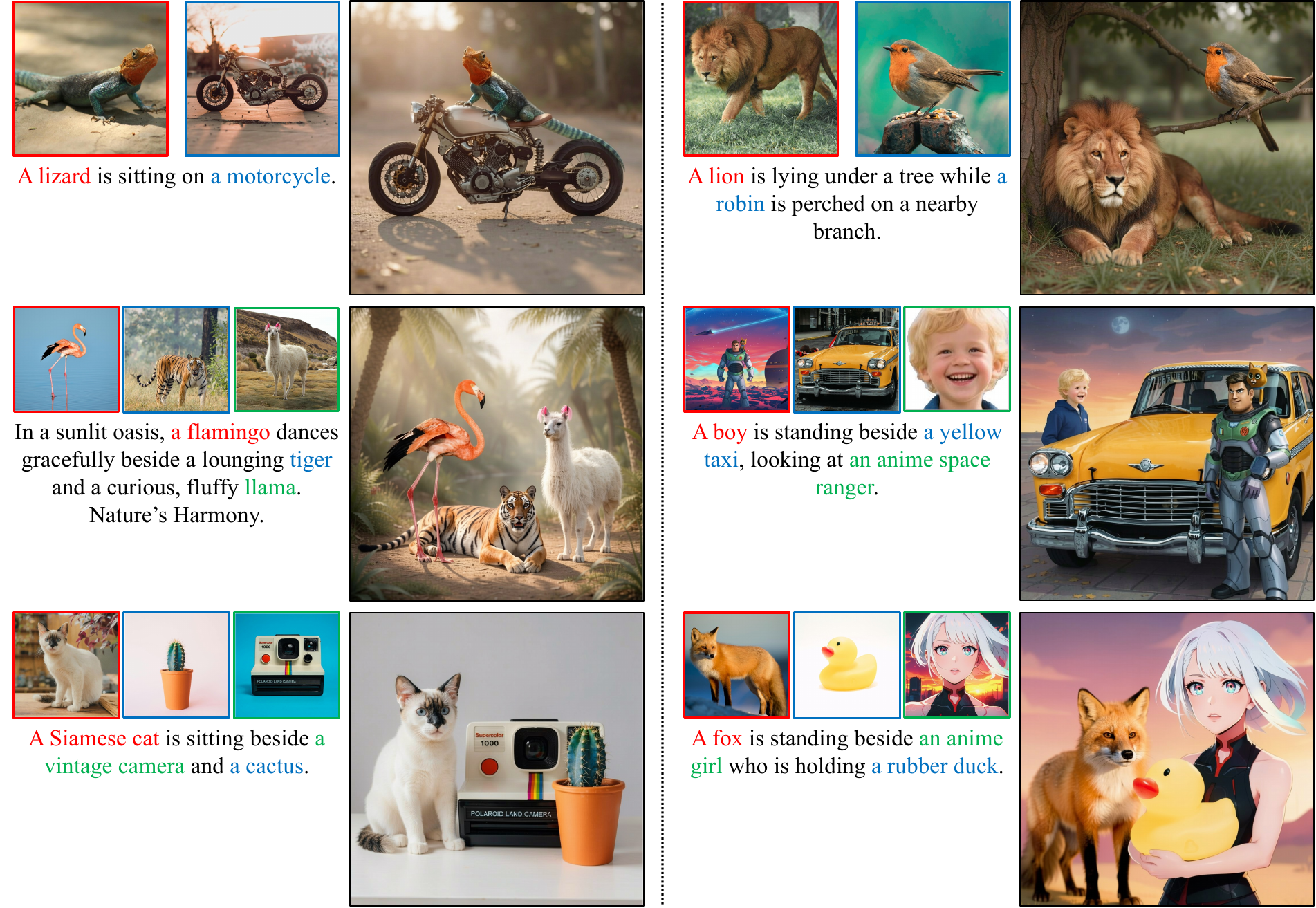}
    \caption{Qualitative results of CRAFT on XVerseBench with un-segmented raw reference images at inference (mask-free protocol).}
    \label{fig:xverse_unseg}
\end{figure*}

\begin{figure*}
    \centering
    \includegraphics[width=.95\linewidth]{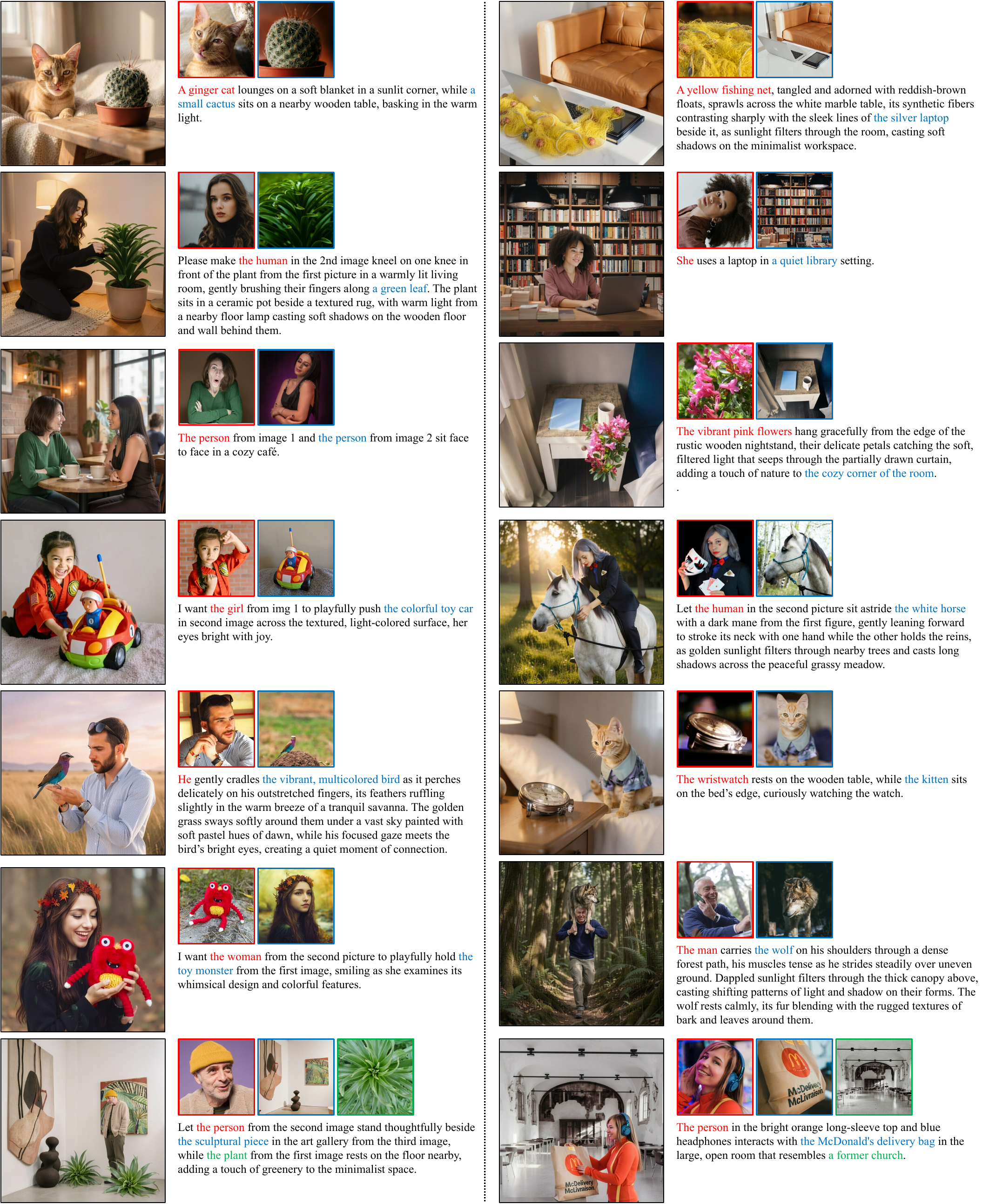}
    \caption{Qualitative results of CRAFT for the \rev{OmniContext} benchmark.}
    \label{fig:omnicontext}
\end{figure*}
\clearpage
\appendix
\clearpage
\appendix
\setcounter{page}{1}
\begin{center}
    \textbf{CRAFT: Constrained Reward via Attention Fine-Tuning for Subject Personalization without Composed Targets --- Supplementary Material}
\end{center}

\section{Subject-Routing Analysis}
\label{app:routing}
Applying ReFL to a large MMDiT's attention map requires selecting a small set of denoising steps and attention blocks on which to place supervision. Reading attention tensors disables Flash Attention \cite{dao2022flashattention, dao2023flashattention2}, and applying uniform supervision across all blocks and steps is computationally inefficient. We therefore perform a lightweight subject-routing analysis on the unmodified backbone and select the step/block coordinates whose noise-to-reference attention best aligns with generated subject masks. 

\noindent\textbf{Measurement protocol.}
We run the unmodified FLUX.2-klein backbone (no LoRA) on $100$ random (reference, prompt) pairs. For each pair, we feed the reference image together with its prompt into the model, obtain one generated image, and compute a Grounded-SAM~\cite{ren2024grounded} mask $\mathbf{m}^\text{gen}$ on the generated image. At every (step, block), we read the noise-to-reference attention $\mathbf{A}_{N2R}$, sum along the reference axis to obtain a noise-grid heatmap, threshold at the $75^\text{th}$ percentile, and compute the binary IoU against $\mathbf{m}^\text{gen}$. A high BinIoU at $(t, b)$ indicates that the base model is already attending to the actual generated subject region at that step and block, so any supervision pulling on this attention amplifies a signal that is already aligned with where the subject ultimately appears. We additionally compute three alternative criteria---binary cross-entropy (BCE), mean squared error (MSE), and soft IoU---on the real-valued (un-thresholded) heatmap, then aggregate by average rank across all four criteria to avoid sensitivity to any single threshold choice.

\noindent\textbf{Step-wise behaviour.}
We sweep all $4 \times 32 = 128$ (step, block) pairs across $100$ subjects. The per-step top-$3$ block mean of $\mathbf{A}_{N2R}$ (noise-to-reference attention) across the four criteria is reported in \tabref{tab:ref_vs_txt_key_per_step}~(b): step $2$ wins on every criterion. The strongest single block is \texttt{single\_1} at $t^\ast{=}2$ (BinIoU $0.666$, see \tabref{tab:ref_vs_txt_key_per_step}~(a)), so $(t^\ast{=}2, \texttt{single\_1})$ is the joint argmax of our sweep.

The per-step \emph{full-network} mean BinIoU is $\{0.334, 0.466, 0.528, 0.549\}$ across steps $\{0, 1, 2, 3\}$: step $3$ has a slightly higher mean across all $32$ blocks, but this average is dominated by $26$ low-alignment blocks whose attention only becomes broadly subject-aware at step $3$. Restricting to the top-$3$ blocks (the ones we would actually supervise) reverses the ranking and places step $2$ as the cleanest reward step.

\noindent\textbf{Block ranking at $t^\ast = 2$ (extended).}
Ranking blocks by averaged rank across the four heatmap-vs-mask criteria, the top three single-stream blocks (\texttt{single\_1}, \texttt{single\_9}, \texttt{single\_8}) are clearly separated from the rest of the network on every individual criterion (BCE, MSE, soft mIoU, BinIoU), not only on the aggregated rank, so the choice of $\mathcal{B}$ is robust to which heatmap-similarity metric is used.

\begin{figure}[t]
    \centering
    \includegraphics[width=\linewidth]{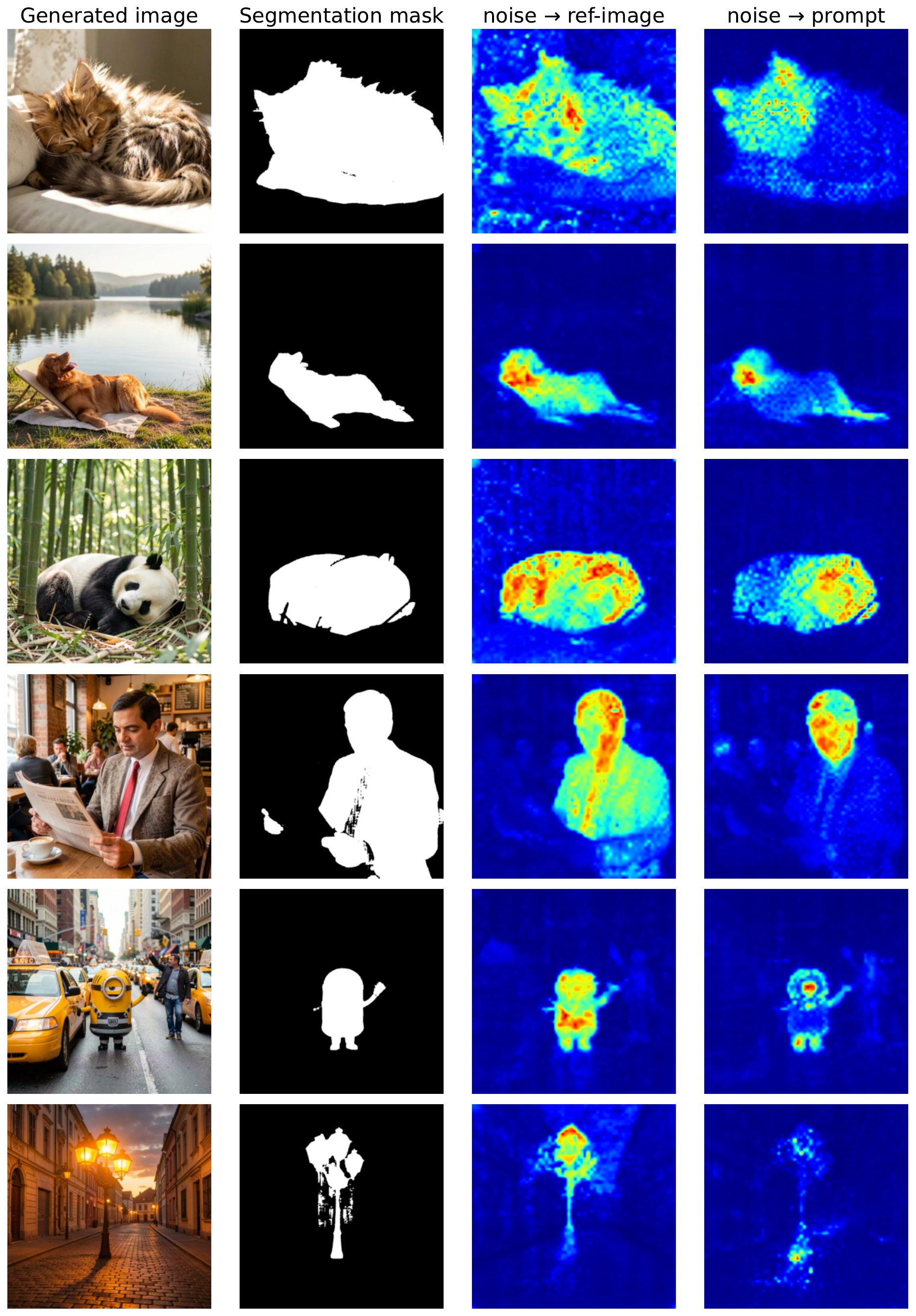}
    \caption{Qualitative comparison of \emph{raw} (un-thresholded) $\mathbf{A}_{N2R}$ (noise-to-reference) and $\mathbf{A}_{N2P}$ (noise-to-prompt) attention heatmaps at $(t^\ast{=}2, \texttt{single\_1})$, shown alongside the Grounded-SAM segmentation of the generated image. $\mathbf{A}_{N2R}$ concentrates on the actual subject region and aligns with the segmentation mask; $\mathbf{A}_{N2P}$ is more diffuse and frequently leaks outside the subject.}
    \label{fig:ref_vs_txt_key_attention}
\end{figure}

\noindent\textbf{Reference-image key vs.\ prompt key.}
We further test whether the reference-image key or the prompt key is the better target for the noise-to-key attention, comparing $\mathbf{A}_{N2R}$ (noise-to-reference) against $\mathbf{A}_{N2P}$ (noise-to-prompt) under the same measurement protocol. \tabref{tab:ref_vs_txt_key_per_step} compares the two against the Grounded-SAM segmentation of the generated image across all four denoising steps, both (a) at the single block \texttt{single\_1} and (b) at the per-step top-$3$ block mean. $\mathbf{A}_{N2R}$ dominates on the majority of metrics across steps, and at our chosen step $t^\ast{=}2$ it wins on every metric in both views. \figref{fig:ref_vs_txt_key_attention} shows the corresponding qualitative comparison on the \emph{raw} (un-thresholded) heatmaps---$\mathbf{A}_{N2R}$ concentrates on the actual subject region in the generated image, while $\mathbf{A}_{N2P}$ is more diffuse and leaks outside the subject. Together, these results support our choice of $\mathbf{A}_{N2R}$ (rather than $\mathbf{A}_{N2P}$) as the primary localization signal that $\mathcal{R}_\text{ref}$ and $\mathbf{m}_k^\text{noise}$ build on.

\begin{table}[t]
\centering\small
\caption{Alignment of attention heatmaps with the Grounded-SAM segmentation of the generated image across all four denoising steps of the $4$-step distilled FLUX.2-klein-9B, on $100$ subjects under the measurement protocol of \secref{app:routing} (BinIoU at the $75$-th percentile binary mask; un-thresholded heatmap for BCE, MSE, soft mIoU; $64{\times}64$ resolution; no Gaussian smoothing). \textbf{Bold}~$=$~better between the two key types within the same step. \textit{(a)} same block \texttt{single\_1}; \textit{(b)} the per-step top-$3$ block mean, where the top-$3$ blocks are selected per step per key type by average rank across the four metrics.}
\label{tab:ref_vs_txt_key_per_step}
\resizebox{.48\textwidth}{!}{%
\begin{tabular}{c|l|cccc}
\toprule
Step & Attention map & BCE $\downarrow$ & MSE $\downarrow$ & soft mIoU $\uparrow$ & BinIoU $\uparrow$ \\
\midrule
\multicolumn{6}{l}{\textit{(a) Same block (\texttt{single\_1}) across steps}} \\
\midrule
$0$ & $\mathbf{A}_{N2R}$ (noise $\to$ ref-image) & 0.327 & \textbf{0.099} & 0.497 & \textbf{0.483} \\
$0$ & $\mathbf{A}_{N2P}$ (noise $\to$ prompt)      & \textbf{0.317} & 0.102 & \textbf{0.503} & 0.440 \\
\midrule
$1$ & $\mathbf{A}_{N2R}$ (noise $\to$ ref-image) & \textbf{0.240} & \textbf{0.071} & \textbf{0.574} & \textbf{0.622} \\
$1$ & $\mathbf{A}_{N2P}$ (noise $\to$ prompt)      & 0.305 & 0.098 & 0.521 & 0.565 \\
\midrule
$\mathbf{2}$ & $\mathbf{A}_{N2R}$ (noise $\to$ ref-image) & \textbf{0.232} & \textbf{0.071} & \textbf{0.586} & \textbf{0.666} \\
$\mathbf{2}$ & $\mathbf{A}_{N2P}$ (noise $\to$ prompt)      & 0.295 & 0.095 & 0.533 & 0.654 \\
\midrule
$3$ & $\mathbf{A}_{N2R}$ (noise $\to$ ref-image) & \textbf{0.246} & \textbf{0.076} & \textbf{0.581} & 0.653 \\
$3$ & $\mathbf{A}_{N2P}$ (noise $\to$ prompt)      & 0.305 & 0.098 & 0.529 & \textbf{0.666} \\
\midrule\midrule
\multicolumn{6}{l}{\textit{(b) Per-step top-$3$ block mean (top-$3$ selected by avg rank)}} \\
\midrule
$0$ & $\mathbf{A}_{N2R}$ (noise $\to$ ref-image) & 0.325 & \textbf{0.098} & 0.492 & \textbf{0.454} \\
$0$ & $\mathbf{A}_{N2P}$ (noise $\to$ prompt)      & \textbf{0.322} & 0.101 & \textbf{0.500} & 0.424 \\
\midrule
$1$ & $\mathbf{A}_{N2R}$ (noise $\to$ ref-image) & \textbf{0.255} & \textbf{0.078} & \textbf{0.555} & \textbf{0.612} \\
$1$ & $\mathbf{A}_{N2P}$ (noise $\to$ prompt)      & 0.281 & 0.088 & 0.534 & 0.545 \\
\midrule
$\mathbf{2}$ & $\mathbf{A}_{N2R}$ (noise $\to$ ref-image) & \textbf{0.243} & \textbf{0.074} & \textbf{0.578} & \textbf{0.656} \\
$\mathbf{2}$ & $\mathbf{A}_{N2P}$ (noise $\to$ prompt)      & 0.274 & 0.087 & 0.545 & 0.638 \\
\midrule
$3$ & $\mathbf{A}_{N2R}$ (noise $\to$ ref-image) & \textbf{0.251} & \textbf{0.077} & \textbf{0.576} & \textbf{0.653} \\
$3$ & $\mathbf{A}_{N2P}$ (noise $\to$ prompt)      & 0.288 & 0.092 & 0.535 & 0.628 \\
\bottomrule
\end{tabular}
}
\end{table}

\begin{figure}
    \centering
    \includegraphics[width=1\linewidth]{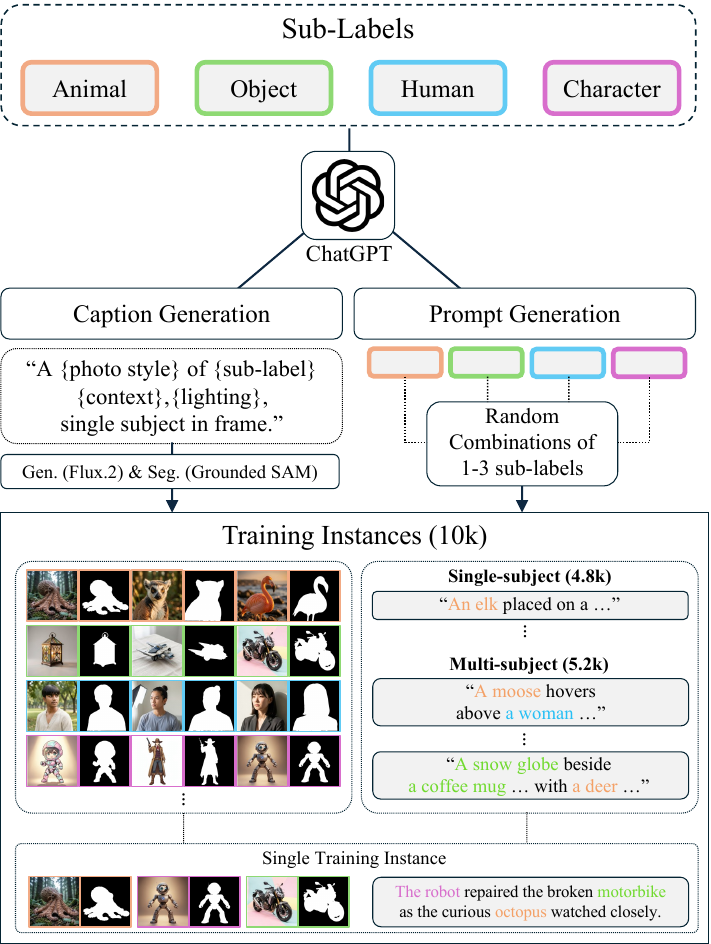}
    \caption{Overall pipeline of dataset curation.}
    \label{fig:dataset_curation}
\end{figure}

\section{Training-Dataset Construction Details}
\label{app:dataset}

Whereas composed-target pipelines synthesize subject-in-scene images and curate them through quality filters and correspondence labeling, our $10{,}000$-instance training dataset is assembled from a single image per subject with no compositional supervision. As illustrated in \figref{fig:dataset_curation}, the pipeline has three lightweight steps: (i) ChatGPT generates short rendering captions over a fixed sub-label vocabulary, (ii) a T2I model renders one isolated image per caption, and Grounded-SAM~\cite{ren2024grounded} extracts a subject mask from each rendered image.  (iii) generate random prompts based on random combinations of 1-3 sub-labels. The resulting reference (image, mask) pairs are then bound to training prompts that mention one or several subjects, yielding single- and multi-subject training instances.

\noindent\textbf{Sub-label vocabulary.}
We list $400$ sub-labels in four top-level categories of $100$ each:

\begin{table*}[t]
\centering\small
\caption{XVerseBench results for the encoder-adapter baselines, with CRAFT included as a reference. Numbers for the encoder-adapter rows are from XVerse~\cite{chen2025xverse}. Best/second-best in \textbf{bold}/\underline{underlined}.}
\label{tab:xverse_encoder_adapter}
\begin{tabular}{l|ccccc|ccccc|c}
\toprule
& \multicolumn{5}{c|}{\textbf{Single subject (90 prompts)}}
& \multicolumn{5}{c|}{\textbf{Multi subject (210 prompts)}}
& \multirow{2}{*}{\textbf{Overall}~$\uparrow$} \\
Method & DPG~$\uparrow$ & ID~$\uparrow$ & IP~$\uparrow$ & AES~$\uparrow$ & AVG~$\uparrow$
       & DPG~$\uparrow$ & ID~$\uparrow$ & IP~$\uparrow$ & AES~$\uparrow$ & AVG~$\uparrow$
       & \\
\midrule
MS-Diffusion~\cite{huang2025msdiffusion} & 96.89 & 6.52  & 55.71 & 59.63 & 54.69 & 87.21 & 3.77  & 46.21 & 55.91 & 48.28 & 51.49 \\
MIP-Adapter~\cite{huang2025mip}          & 87.56 & 39.59 & 71.97 & 52.12 & 62.81 & 84.56 & 24.58 & 57.00 & 51.81 & 54.49 & 58.65 \\
\midrule
\textbf{CRAFT (Ours)}                    & 96.81 & 84.22 & 84.23 & 61.24 & 81.62 & 88.71 & 61.16 & 77.25 & 58.16 & 71.32 & 76.47 \\
\bottomrule
\end{tabular}%

\end{table*}

\begin{itemize}
\setlength\itemsep{2pt}
\item \textbf{Animal:} mammals, birds, reptiles, amphibians, marine life, insects.
\item \textbf{Object:} furniture, electronics, vehicles, food, fashion, tools, decor, plants, toys.
\item \textbf{Human:} sampled from $21$ ethnicities $\times$ four age groups $\times$ two genders, with optional profession or attire.
\item \textbf{Character:} stylized archetypes (anime, mecha, cyberpunk, Pixar 3D, chibi, low-poly, comic, steampunk, etc.).
\end{itemize}
\noindent\textbf{Rendering reference images.}
For each sub-label, ChatGPT generates $12$ short captions varying in (context, lighting, camera angle, photo style) under a ``single subject in frame'' constraint, yielding $4{,}800$ captions in total. The captions are tightly templated per category to keep the photo compositionally clean: animal captions request a single creature in soft natural light, object captions place the item in a neutral interior at a fixed three-quarter angle, human captions request a head-and-shoulders portrait, and character captions request a full-body stylized illustration. For example, an animal caption reads ``\emph{A photorealistic close-up of a crow in a meadow filled with wildflowers, soft diffuse natural light, single subject in frame.}'' Each caption is rendered once at $1024^2$ with FLUX.2-klein, and Grounded-SAM~\cite{ren2024grounded} then extracts a subject mask, which we downsample to $64^2$ to match the noise-grid resolution. No quality filtering is applied; each (image, mask) pair is kept as produced.

\noindent\textbf{Training prompts.}
Training prompts are also generated by ChatGPT and cover all combinations of $2$ to $3$ subjects drawn from the four categories (animal, object, human, character), with category repetition allowed (e.g., two different objects in the same scene). Each prompt is a free-form scene with one to three sub-labels dropped in. For example: ``An \textbf{elk} placed on a statue of a philosopher in a sunlit classical garden'' (single-subject, animal); ``a floating \textbf{moose} hovers above a glowing \textbf{Polynesian middle-aged woman}, suspended in a twilight sky'' (multi-subject, animal~$+$~human); or ``a \textbf{snow globe} beside a \textbf{coffee mug with latte art} on a wooden table by the window, with a \textbf{white-tailed deer} curled up nearby'' (multi-subject, animal~$+$~object~$+$~object). Each filled prompt is paired with the corresponding reference (image, mask) tuples to form a training instance. Crucially, no training prompt is ever rendered into a composed-target image; the text alone is supplied to the model during training, and the rewards (\secref{sec:method}) provide the only signal that the generated output should respect both the prompt and the references. The final dataset has $4{,}800$ single-subject and $5{,}200$ multi-subject instances ($10{,}000$ total).

\section{Encoder-Adapter Baselines on XVerseBench}
\label{app:encoder_adapter}

We report the two encoder-adapter baselines---MS-Diffusion~\cite{huang2025msdiffusion} and MIP-Adapter~\cite{huang2025mip}---deferred from the models in \secref{sec:exp:comparison}. Both methods inject reference features at inference time without per-subject fine-tuning, and were originally trained on category-specific subjects (e.g., faces, objects). \tabref{tab:xverse_encoder_adapter} reports their XVerseBench scores together with CRAFT for reference; numbers are taken from XVerse~\cite{chen2025xverse} (Tab.~1).

The dominant gap is on identity preservation: MS-Diffusion drops to ID $6.52$ on single-subject and $3.77$ on multi-subject, and MIP-Adapter to $39.59$ and $24.58$, reflecting their specialization to specific subject categories at training time. CRAFT exceeds both baselines on every per-metric column on both splits, with the sole exception of Single DPG where MS-Diffusion is marginally higher ($96.89$ vs.\ $96.81$), supporting the observation that encoder-adapter approaches transfer poorly to the diverse subject distribution of XVerseBench.

\section{Generalization Across Reference-Aware MMDiT Backbones}
\label{app:uno}

To verify that CRAFT is not specific to FLUX.2-klein, we apply the same reward design to two additional reference-aware MMDiT backbones built on FLUX.1-dev: (i) the UNO~\cite{wu2025uno} reference adapter, and (ii) UMO~\cite{cheng2025umo}, itself a reward fine-tune of UNO. In both cases, we add a CRAFT LoRA on top of the existing (FLUX.1-dev~$+$~base LoRA), training only the new adapter weights. The headline results appear in \tabref{tab:uno_craft} (\secref{sec:exp:other-backbone}). This appendix provides the per-metric breakdown and implementation details.

\noindent\textbf{Reward step and block selection.}
The UNO/UMO backbones use the standard FLUX.1-dev sampling schedule with $T = 25$ flow-matching steps, in contrast to the $4$-step distilled schedule of FLUX.2-klein. Because the longer schedule provides room for window-based reward sampling, we adopt the original ReFL recipe~\cite{xu2024imagereward} as-is and replace the fixed reward step $t^\ast$ with a uniform sample $t \sim \mathcal{U}[T_S, T_E]$ over $[T_S, T_E] = [10, 15]$, corresponding roughly to the $40$--$60\%$ window of the schedule's progress. For the reward blocks $\mathcal{B}$, we follow the routing-analysis protocol of \secref{app:routing} on the unmodified UNO backbone and select the three single-stream blocks with the highest BinIoU in this window, giving $\mathcal{B} = \{\texttt{single\_23}, \texttt{single\_25}, \texttt{single\_26}\}$ for UNO; the same $\mathcal{B}$ is reused for UMO+CRAFT.

\noindent\textbf{Training configuration.}
For each backbone, we attach a rank-$512$ LoRA matching the existing UNO/UMO LoRA shape; for UMO+CRA\\FT, the LoRA is initialized from the released UMO\_UNO weights so that CRAFT trains a delta on top of UMO's identity-tuned state. We train at $768^2$ resolution with the $25$-step schedule above, learning rate $5{\times}10^{-6}$, $w_\text{anchor}{=}0.3$, $w_\text{a}{=}5{\times}10^{-3}$, $w_\text{nr} = w_\text{tr} = 0.5$ (inside $\mathcal{R}_\text{ref}$), $w_\text{c}{=}0.3$, $w_\text{id}{=}3.0$. We use the same training dataset as the main experiments. Each variant trains for $3{,}000$ steps on $4$ B200 GPUs ($\sim\!4$ hours per run). The full procedure is summarized in \algref{alg:craft_uno}; compared to \algref{alg:craft} in the main paper, the only difference is that the reward step is sampled from a window $[T_S, T_E]$ following the original ReFL recipe~\cite{xu2024imagereward}, rather than fixed at $t^\ast$.

\begin{table*}[t]
\centering\small
\caption{Per-metric breakdown of \tabref{tab:uno_craft}: CRAFT applied to UNO~\cite{wu2025uno} and UMO~\cite{cheng2025umo} on XVerseBench. Best in each column in \textbf{bold}.}
\label{tab:uno_xversebench}
\begin{tabular}{l|ccccc|ccccc|c}
\toprule
& \multicolumn{5}{c|}{\textbf{Single subject (90 prompts)}}
& \multicolumn{5}{c|}{\textbf{Multi subject (210 prompts)}}
& \multirow{2}{*}{\textbf{Overall}~$\uparrow$} \\
Method & DPG~$\uparrow$ & ID~$\uparrow$ & IP~$\uparrow$ & AES~$\uparrow$ & AVG~$\uparrow$
       & DPG~$\uparrow$ & ID~$\uparrow$ & IP~$\uparrow$ & AES~$\uparrow$ & AVG~$\uparrow$
       & \\
\midrule
UNO~\cite{wu2025uno}                   & \textbf{89.65} & 47.91 & 80.40 & 55.90 & 68.47 & 85.28 & 31.82 & 67.00 & 54.24 & 59.59 & 64.03 \\
\textbf{UNO + CRAFT (Ours)}                        & 85.49 & 74.43 & 84.90 & 61.60 & 76.61 & 86.37 & 49.60 & \textbf{74.14} & 58.04 & 67.04 & 71.83 \\
\midrule
UMO~\cite{cheng2025umo} & 86.75 & 77.36 & 76.99 & \textbf{61.71} & 75.70 & \textbf{87.18} & 58.24 & 60.52 & 58.74 & 66.17 & 70.94 \\
\textbf{UMO + CRAFT (Ours)}                        & 85.92 & \textbf{86.26} & \textbf{85.51} & 61.32 & \textbf{79.75} & 86.45 & \textbf{62.36} & 70.10 & \textbf{59.05} & \textbf{69.49} & \textbf{74.62} \\
\bottomrule
\end{tabular}
\end{table*}

\begin{algorithm}[t]
\caption{CRAFT training on the UNO/UMO backbone (single-step ReFL).}
\label{alg:craft_uno}
\begin{algorithmic}[1]
\REQUIRE Base model $f_\text{base}$ (frozen; FLUX.1-dev~$+$~UNO LoRA, or FLUX.1-dev~$+$~UMO LoRA) and CRAFT-LoRA-adapted model $f_\text{LoRA}$ with learnable parameters $\Delta\theta$, total denoising steps $T = 25$, \textbf{reward step window} $[T_S, T_E] = [10, 15]$, reward blocks $\mathcal{B} = \{\texttt{single\_23}, \texttt{single\_25}, \texttt{single\_26}\}$, learning rate $\eta = 5{\times}10^{-6}$.
\FOR{each training instance $(y, \{\mathbf{I}_k^\text{ref}, \mathbf{M}_k^\text{ref}\}_{k=1}^{K})$}
    \STATE Sample $\mathbf{z}_T \sim \mathcal{N}(\mathbf{0}, \mathbf{I})$ and $t \sim \mathcal{U}[T_S, T_E]$. \hfill \textcolor{blue}{// reward step within the window}
    \FOR{$\tau = T, T{-}1, \dots, t{+}1$}
        \STATE \textbf{no grad:} $\mathbf{z}_{\tau-1} \gets \mathbf{z}_\tau - \tau\, f_\text{base}(\mathbf{z}_\tau, \tau \mid y, \{\mathbf{I}_k^\text{ref}\})$ \hfill \textcolor{blue}{// roll out $T{-}t$ denoising steps}
    \ENDFOR
    \STATE \textbf{no grad:} $\mathbf{v}_\text{base} \gets f_\text{base}(\mathbf{z}_t, t \mid y, \{\mathbf{I}_k^\text{ref}\})$ \hfill \textcolor{blue}{// base velocity}
    \STATE \textbf{with grad:} $\mathbf{v}_\text{LoRA},\, \mathcal{A} \gets f_\text{LoRA}(\mathbf{z}_{t^\ast}, t^\ast \mid y, \{\mathbf{I}^\text{ref}_k\})$ \hfill \textcolor{blue}{// $\mathcal{A} = \{\mathbf{A}_{N2R_k}, \mathbf{A}_{N2P_k}, \mathbf{A}_{P_k 2 R_k}\}_{k=1}^{K}$}
    \STATE Decode the pre-image: $\hat{\mathbf{I}} \gets \mathcal{D}_\text{VAE}(\mathbf{z}_t - t\, \mathbf{v}_\text{LoRA})$.
    \STATE Compute attention rewards $\mathcal{R}_\text{ref}, \mathcal{R}_\text{cons}$ and per-subject masks $\{\mathbf{m}^\text{noise}_k\}$ via Equations~\eqref{eq:r_noise_ref}-\eqref{eq:m_attn}.
    \STATE Compute identity reward $\mathcal{R}_\text{id}$ via \equref{eq:r_id}; auxiliary terms $\mathcal{R}_\text{CLIP-T}, \mathcal{R}_\text{AES}, \mathcal{L}_\text{anchor}$ as defined in \secref{sec:method:loss}.
    \STATE Form the total loss $\mathcal{L}$ via \equref{eq:total_loss} and update $\Delta\theta \gets \Delta\theta - \eta\, \nabla_{\Delta\theta}\, \mathcal{L}$.
\ENDFOR
\end{algorithmic}
\end{algorithm}

\noindent\textbf{Discussion.} 
As shown in \tabref{tab:uno_xversebench}, CRAFT consistently improves over both backbones. Against the UNO base, CRAFT lifts single-subject AVG from $68.47$ to $76.61$ ($+8.14$) and multi-subject AVG from $59.59$ to $67.04$ ($+7.45$), for $+7.80$ overall. The dominant gains are on identity (Single ID $47.91 \to 74.43$, $+26.52$; Multi ID $31.82 \to 49.60$, $+17.78$) and IP (Single $80.40 \to 84.90$, $+4.50$; Multi $67.00 \to 74.14$, $+7.14$), with AES rising on both splits ($+5.70$ Single, $+3.80$ Multi). DPG stays stable on the multi-subject split ($+1.09$) but drops on the single-subject split ($89.65 \to 85.49$, $-4.16$), reflecting a small trade-off between sharper identity routing and prompt-element coverage.

Composed on top of UMO's identity-tuned LoRA, CRAFT raises single AVG from $75.70$ to $79.75$ ($+4.05$) and multi AVG from $66.17$ to $69.49$ ($+3.32$), for $+3.68$ overall, with only a small drop on UMO's multi-subject DPG ($86.45$ vs.\ $87.18$, $-0.73$). UMO+CRAFT is the best method on the Single AVG, Multi AVG, and Overall aggregates, and on Single ID, Single IP, Multi ID, and Multi AES. The remaining columns are split: UNO base wins Single DPG, UMO base wins Single AES and Multi DPG, and UNO+CRAFT wins Multi IP. This composability behavior---CRAFT improving even on top of an already identity-tuned LoRA---supports the interpretation of CRAFT as a complementary attention-grounded reward stack rather than a competing identity reward.

\section{Additional Personalization Benchmarks}
\label{app:additional_benchmarks}

Beyond XVerseBench (\secref{sec:experiment}), we evaluate CRAFT on two additional benchmarks: the OmniContext~\cite{wu2025omnigen2} for personalization and composition (\secref{app:omnicontext}), and DreamBench~\cite{ruiz2023dreambooth} for single-subject personalization (\secref{app:dreambench}).

\subsection{OmniContext: Personalization-and-Composition Evaluation}
\label{app:omnicontext}

\noindent\textbf{Benchmark.}
OmniContext~\cite{wu2025omnigen2}, released with OmniGen2, targets free-form scene composition rather than the canonical, single-image personalization setting of XVerseBench. It contains $400$ instructions evenly split across eight task types: \emph{Single} (Char., Obj.), \emph{Multiple} (Char., Obj., Char.~$+$~Obj.), and \emph{Scene} (Char., Obj., Char.~$+$~Obj.), where ``Scene'' tasks additionally require placing the subject into a separately specified background image. Each instruction is paired with one to three reference images and an open-vocabulary natural-language editing/composition prompt.

\noindent\textbf{Evaluation Metrics.}
We follow the OmniContext protocol~\cite{wu2025omnigen2} and report two GPT-judged scores on a $0$--$10$ scale: \emph{Prompt Following} (PF), which measures whether the generated image satisfies the textual instruction, and \emph{Subject Consistency} (SC), which measures whether reference subjects retain their identity in the output. The two scores are averaged per task type, then macro-averaged across the eight task types to give the overall AVG. Numbers for prior methods are taken verbatim from Scone~\cite{wang2025scone} and OmniGen2~\cite{wu2025omnigen2}; CRAFT and the FLUX.2-klein backbone are evaluated under the same protocol with our own GPT-4.1 judge using the same scoring template.

\noindent\textbf{Discussion.}
\tabref{tab:omnicontext} shows that CRAFT is the strongest open-source method on OmniContext, leading every column and improving the overall AVG by $+0.71$ over the previous open-source best, Scone~\cite{wang2025scone} ($8.72$ vs.\ $8.01$). The largest gaps over Scone are on the \emph{Single} tasks (Single Char.\ $+0.71$, Single Obj.\ $+0.77$) and on the \emph{Scene} tasks (Scene Char.\ $+1.52$, Scene Char.~$+$~Obj.\ $+0.57$), which together account for the bulk of the AVG improvement. Against the FLUX.2-klein backbone ($8.62$), CRAFT improves overall by $+0.10$ AVG, with the largest per-task gains on the multi-subject splits (Multi Char.\ $+0.34$, Multi C.~$+$~O.\ $+0.16$, Multi Obj.\ $+0.14$)---the splits where attention-routing supervision is most informative. The remaining gap to GPT-4o ($8.78$) is only $0.06$ AVG and is concentrated on the \emph{Scene} tasks, where the closed-source system still benefits from a much larger image-generation backbone. Qualitative examples are shown in \figref{fig:omnicontext}.

\begin{table*}[t]
\centering\small
\caption{Quantitative comparison on OmniContext~\cite{wu2025omnigen2}. ``Char.~$+$~Obj.''~indicates Character~$+$~Object. Methods are partitioned into closed-source proprietary and open-source systems following Scone~\cite{wang2025scone}; numbers for prior methods are from Scone~\cite{wang2025scone} and OmniGen2~\cite{wu2025omnigen2}, while CRAFT and the FLUX.2-klein backbone are evaluated under the same protocol. The FLUX.2-klein backbone is shown as a reference and is not included in the ranking. Best/second-best within each group in \textbf{bold}/\underline{underlined}.}
\label{tab:omnicontext}
\begin{tabular}{lccccccccc}
\toprule
& \multicolumn{2}{c}{\textbf{Single}} & \multicolumn{3}{c}{\textbf{Multiple}} & \multicolumn{3}{c}{\textbf{Scene}} & \\
\cmidrule(lr){2-3}\cmidrule(lr){4-6}\cmidrule(lr){7-9}
Method & Char.~$\uparrow$ & Obj.~$\uparrow$ & Char.~$\uparrow$ & Obj.~$\uparrow$ & C.~$+$~O.~$\uparrow$ & Char.~$\uparrow$ & Obj.~$\uparrow$ & C.~$+$~O.~$\uparrow$ & \textbf{AVG~$\uparrow$} \\
\midrule
\multicolumn{10}{l}{\textit{Closed-source proprietary}} \\
\midrule
GPT-4o                            & \textbf{8.96} & \underline{8.91} & \textbf{8.90} & \textbf{8.95} & \textbf{8.81} & \textbf{8.92} & \textbf{8.40} & \textbf{8.44} & \textbf{8.78} \\
Gemini-2.5-Flash-Image            & \underline{8.79} & \textbf{9.12} & \underline{8.27} & \underline{8.60} & \underline{7.71} & \underline{7.63} & \underline{7.65} & \underline{6.81} & \underline{8.07} \\
\midrule
\multicolumn{10}{l}{\textit{Open-source}} \\
\midrule
FLUX.2-klein~\cite{flux2} (base)  & 8.92 & 9.20 & 8.56 & 8.75 & 8.52 & 8.51 & 8.24 & 8.25 & 8.62 \\
\midrule
UNO~\cite{wu2025uno}              & 7.15 & 6.72 & 3.56 & 6.46 & 4.90 & 2.72 & 4.89 & 4.76 & 5.14 \\
USO~\cite{wu2025uso}                                & 8.03 & 7.55 & 3.32 & 6.10 & 4.56 & 2.77 & 5.38 & 5.09 & 5.35 \\
BAGEL~\cite{deng2025emerging}                              & 7.00 & 7.04 & 5.32 & 6.69 & 6.74 & 3.94 & 5.77 & 5.73 & 6.03 \\
OmniGen2~\cite{wu2025omnigen2}    & 8.17 & 7.63 & 7.26 & 7.03 & 7.56 & 7.02 & 6.90 & 6.64 & 7.28 \\
UniWorld-V2~\cite{li2025uniworld}                        & 8.45 & 8.44 & 7.87 & 8.22 & 7.95 & 5.36 & 7.47 & 6.98 & 7.59 \\
Qwen-Image-Edit-2509~\cite{wu2025qwenimagetechnicalreport}               & \underline{8.56} & 8.41 & 7.92 & \underline{8.37} & 7.79 & 5.23 & 7.70 & 6.86 & 7.60 \\
Echo-4o~\cite{ye2025echo}                            & 8.34 & 8.27 & 8.13 & 8.14 & 8.11 & \underline{7.07} & 7.73 & \underline{7.77} & 7.95 \\
Scone~\cite{wang2025scone}        & 8.34 & \underline{8.52} & \underline{8.24} & 8.14 & \underline{8.30} & 7.06 & \underline{7.88} & 7.63 & \underline{8.01} \\
\textbf{CRAFT (Ours)}              & \textbf{9.05} & \textbf{9.29} & \textbf{8.90} & \textbf{8.89} & \textbf{8.68} & \textbf{8.58} & \textbf{8.20} & \textbf{8.20} & \textbf{8.72} \\
\bottomrule
\end{tabular}%
\end{table*}

\begin{table*}[t]
\centering\small
\caption{Effect of training-set size on XVerseBench. The recipe is retrained on uniformly-sampled subsets of the $10{,}000$-instance training dataset (\secref{app:dataset}); all other hyperparameters are unchanged. $\Delta$ is the change in Overall relative to the full $10{,}000$-instance baseline used in the main paper. \rev{The $10{,}000$ row (\textbf{bold}) is the configuration used in the main paper.} \rev{The $20{,}000$ row extends the corpus with the same recipe (a superset of the $10{,}000$ set) and is shown for reference.}}
\label{tab:dataset_size}
\begin{tabular}{r|ccccc|ccccc|cc}
\toprule
& \multicolumn{5}{c|}{\textbf{Single subject}}
& \multicolumn{5}{c|}{\textbf{Multi subject}}
& \multicolumn{2}{c}{} \\
\#Data & DPG~$\uparrow$ & ID~$\uparrow$ & IP~$\uparrow$ & AES~$\uparrow$ & AVG~$\uparrow$
        & DPG~$\uparrow$ & ID~$\uparrow$ & IP~$\uparrow$ & AES~$\uparrow$ & AVG~$\uparrow$
        & Overall~$\uparrow$ & $\Delta$ \\
\midrule
\rev{$20{,}000$} & \rev{$97.78$} & \rev{$84.09$} & \rev{$84.09$} & \rev{$61.34$} & \rev{$81.83$} & \rev{$87.85$} & \rev{$62.73$} & \rev{$77.23$} & \rev{$58.52$} & \rev{$71.58$} & \rev{$76.70$} & \rev{$+0.23$} \\
$10{,}000$ & \textbf{96.81} & \textbf{84.22} & \textbf{84.23} & \textbf{61.24} & \textbf{81.62} & \textbf{88.71} & \textbf{61.16} & \textbf{77.25} & \textbf{58.16} & \textbf{71.32} & \textbf{76.47} & --- \\
\midrule
$5{,}000$  & 96.94 & 84.78 & 79.90 & 60.95 & 80.64 & 88.15 & 61.43 & 77.40 & 57.81 & 71.20 & 75.92 & $-0.55$ \\
$2{,}000$  & 96.94 & 84.44 & 80.73 & 60.62 & 80.68 & 88.37 & 60.80 & 76.72 & 57.70 & 70.90 & 75.79 & $-0.68$ \\
$1{,}000$  & 96.11 & 82.10 & 81.76 & 61.02 & 80.25 & 86.45 & 59.27 & 76.45 & 59.18 & 70.34 & 75.30 & $-1.18$ \\
\bottomrule
\end{tabular}
\end{table*}

\subsection{DreamBench: Single-Subject Evaluation}
\label{app:dreambench}

\noindent\textbf{Benchmark.}
DreamBench~\cite{ruiz2023dreambooth} contains $30$ subjects (objects, plushies, animals) with $25$ prompts each, for $750$ (subject, prompt) pairs. For each pair, we use the supplied canonical reference image for the subject and generate a single image with our model.

\noindent\textbf{Evaluation Metrics.}
Following recent DreamBench-evaluated baselines~\cite{wu2025uno, mou2025dreamo, chen2025xverse, she2025mosaic} and the official \texttt{dreambench\_plus} evaluator~\cite{peng2025dreambenchpp}, we report three metrics. To assess overall subject-appearance preservation, we compute the \emph{CLIP image similarity}, denoted CLIP-I, as the cosine similarity between CLIP-ViT-B/32~\cite{radford2021learning} embeddings of the reference and generated images. To assess prompt fidelity, we compute the \emph{CLIP text--image similarity}, denoted CLIP-T, between the embedding of the textual prompt and that of the generated image under the same CLIP encoder. To assess fine-grained identity preservation, we compute the \emph{DINO similarity}, denoted DINO, between DINOv2~\cite{oquab2024dinov2} features of the reference and generated images.

\tabref{tab:dreambench} shows that CRAFT is best on CLIP-I and DINO. The CLIP-I gain ($+0.70$ over MOSAIC) and the DINO gain ($+0.66$ over MOSAIC, $+2.62$ over XVerse, $+2.03$ over DreamO) point to the same identity-preservation strength we observe on XVerseBench's ID column: by attending tightly to the reference subject region, CRAFT recovers more of the reference's fine appearance than the encoder-based or composed-target baselines. CLIP-T is within $0.34$ of the best (MOSAIC $31.64$), so the identity gain is not bought at the cost of prompt alignment.

\begin{table}[t]
\centering\small
\caption{Quantitative comparison on the single-subject DreamBench~\cite{ruiz2023dreambooth} benchmark ($30$ subjects, $25$ prompts each). All scores are computed by the official \texttt{dreambench\_plus} evaluator. Best in each column in \textbf{bold}.}
\label{tab:dreambench}
\setlength{\tabcolsep}{6pt}
\begin{tabular}{l|ccc}
\toprule
Method & CLIP-I~$\uparrow$ & CLIP-T~$\uparrow$ & DINO~$\uparrow$ \\
\midrule
DreamBooth~\cite{ruiz2023dreambooth}              & 80.30 & 30.52 & 66.81 \\
BLIP-Diffusion~\cite{li2024blip}                   & 80.47 & 30.24 & 69.82 \\
SSR-Encoder\rev{~\cite{zhang2024ssr}}                                        & 82.10 & 30.79 & 61.22 \\
MS-Diffusion~\cite{huang2025msdiffusion}           & 80.82 & 31.05 & 70.32 \\
UNO~\cite{wu2025uno}                               & 83.50 & 30.41 & 75.97 \\
DreamO~\cite{mou2025dreamo}                        & 83.35 & 30.61 & 76.03 \\
XVerse~\cite{chen2025xverse}                       & 83.20 & 30.20 & 75.44 \\
MOSAIC~\cite{she2025mosaic}                        & 84.30 & \textbf{31.64} & 77.40 \\
\midrule
\textbf{CRAFT (Ours)}                              & \textbf{85.00} & 31.30 & \textbf{78.06} \\
\bottomrule
\end{tabular}
\end{table}

\section{Mask-Free Inference Evaluation}
\label{app:mask_free}

The masks used during training (\secref{sec:method}) are reward-side annotations only; CRAFT itself accepts an un-masked reference at inference. To validate that the trained model is robust without segmentation at test time, we re-evaluate on XVerseBench by supplying the raw (un-segmented) reference images directly to CRAFT, while all other settings follow the official protocol of \secref{sec:exp:eval}. The quantitative comparison is reported in \tabref{tab:xversebench} of the main paper.

The mask-free protocol matches or exceeds the segmented-input setting on every column except Multi ID (where it trails by $0.81$), and improves Overall by $+1.33$ points. The single-subject split benefits the most ($+2.35$ on Single AVG), suggesting that supplying a segmented reference does not provide information that CRAFT cannot recover from the raw image at inference. This confirms that the masks used during training serve only as reward-side annotations, and the deployed model does not depend on segmentation at test time. Qualitative results under both protocols are shown in \figref{fig:xverse_seg} and \figref{fig:xverse_unseg}.

\begin{figure*}
    \centering
    \includegraphics[width=1\linewidth]{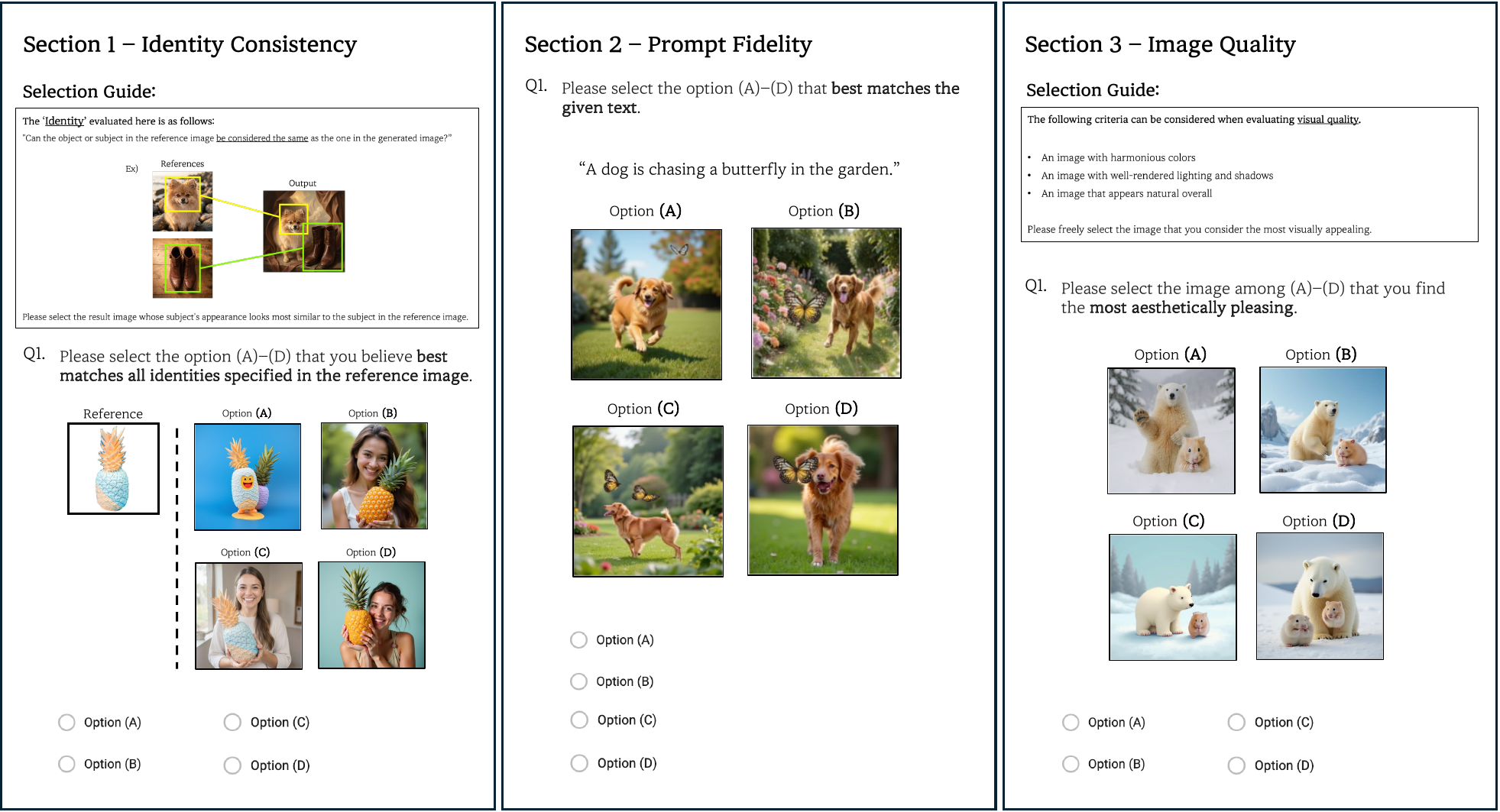}
    \caption{Example interface used in the user study. Participants selected the best-performing method among four candidates for each question.}
    \label{fig:user_study_interface}
\end{figure*}

\section{User Study Details}
\label{app:user_study_details}

To complement the quantitative evaluation, we conducted a user study with $119$ participants aged between $20$ and $50$. Each participant was shown a text prompt together with four sets of generated images (see \figref{fig:user_study_interface} for an example survey screen), each set corresponding to one of the four methods compared in \tabref{tab:user_study}: our CRAFT, UMO~\cite{cheng2025umo}, XVerse~\cite{chen2025xverse}, and MOSAIC~\cite{she2025mosaic}. Participants were instructed to select the image set that best satisfied each of the following criteria:

\begin{itemize}
\item \textbf{Identity Consistency (IC):} Please select the option (A)–(D) that you believe best matches all identities specified in the reference image.
\item \textbf{Prompt Fidelity (PF):} Please select the option (A)–(D) that best matches the given text.
\item \textbf{Image Quality (IQ):} Please select the image among (A)–(D) that you find the most aesthetically pleasing.
\end{itemize}

Each participant evaluated multiple sets across diverse prompts in a randomized order.

\section{Training-Set Size Ablation}
\label{app:dataset_size}

We assess how much of CRAFT's performance depends on the size of the training dataset by retraining the same recipe on uniformly sampled subsets of the $10{,}000$-instance dataset described in \secref{app:dataset}. Subsets are drawn without replacement with a fixed seed so that smaller subsets are nested in the larger ones, and the single-/multi-subject ratio is preserved within $\pm 0.1\%$ across sizes.

\noindent\textbf{Setup.}
All other hyperparameters match the main run (\secref{sec:exp_impl}).

\noindent\textbf{Results.}
\tabref{tab:dataset_size} reports the per-metric and aggregate scores. Overall improves monotonically with training-set size, but the trend is shallow: scaling data $10\times$ from $1{,}000$ to $10{,}000$ raises Overall by only $1.18$ points ($75.30 \to 76.47$). \rev{Extending in the opposite direction, doubling the corpus to $20{,}000$ instances with the same recipe and matched per-sample exposure adds only $+0.23$ Overall ($\to 76.70$), confirming that $10{,}000$ is already near the plateau.} The per-subject averages show the same pattern, with Single AVG and Multi AVG both gaining around one point over the same range. This shallow scaling suggests that the bulk of CRAFT's gains come from per-instance attention-routing supervision rather than from the size of the training dataset, consistent with the framing of CRAFT as shaping pre-existing attention patterns rather than learning subject features from data.

\begin{table*}[t]
\centering\small
\caption{Reward-weight sensitivity on XVerseBench. Each row perturbs a single weight by $0.5\times$ or $2\times$ relative to the main configuration ($w_\text{nr} = w_\text{tr} = 0.5$, $w_\text{c} = 1.0$, $w_\text{id} = 1.0$); all other weights and training hyperparameters are unchanged. $\Delta$ is the change in Overall vs.\ the baseline row. ``MR'' is the mean rank across the eight raw per-metric columns (Single/Multi $\times$ DPG, ID, IP, AES); lower is more balanced. \textbf{Bold} marks the strict best Overall and the strict lowest mean rank; the baseline row is shown in italics.}
\label{tab:reward_weight_sensitivity}
\resizebox{1\textwidth}{!}{%
\begin{tabular}{l|ccccc|ccccc|ccc}
\toprule
& \multicolumn{5}{c|}{\textbf{Single subject}}
& \multicolumn{5}{c|}{\textbf{Multi subject}}
& \multicolumn{3}{c}{} \\
Variant & DPG~$\uparrow$ & ID~$\uparrow$ & IP~$\uparrow$ & AES~$\uparrow$ & AVG~$\uparrow$
        & DPG~$\uparrow$ & ID~$\uparrow$ & IP~$\uparrow$ & AES~$\uparrow$ & AVG~$\uparrow$
        & Overall~$\uparrow$ & $\Delta$ & MR~$\downarrow$ \\
\midrule
\textit{Baseline} ($w_\text{id}=1.0$, $w_\text{c}=1.0$, $w_\text{nr}=w_\text{tr}=0.5$)
                                  & 96.81 & 84.22 & 84.23 & 61.24 & 81.62 & 88.71 & 61.16 & 77.25 & 58.16 & 71.32 & 76.47 & --- & \textbf{2.62} \\
\midrule
$w_\text{nr} = w_\text{tr} = 1.0$ ($2\times$) & 97.22 & 84.68 & 83.86 & 60.62 & 81.59 & 88.21 & 61.47 & 77.08 & 58.13 & 71.22 & 76.40 & $-0.07$ & 3.12 \\
$w_\text{c} = 0.5$ ($0.5\times$)             & 97.78 & 82.20 & 81.43 & 60.82 & 80.56 & 88.67 & 59.78 & 76.95 & 58.40 & 70.95 & 75.76 & $-0.71$ & 3.50 \\
$w_\text{c} = 2.0$ ($2\times$)               & 97.33 & 83.83 & 83.21 & 60.95 & 81.33 & 85.98 & 60.51 & 76.25 & 57.62 & 70.09 & 75.71 & $-0.76$ & 4.25 \\
$w_\text{id} = 0.5$ ($0.5\times$)            & 97.50 & 81.33 & 83.51 & 61.90 & 81.06 & 87.50 & 57.61 & 75.97 & 59.21 & 70.07 & 75.57 & $-0.90$ & 3.62 \\
$w_\text{id} = 2.0$ ($2\times$)              & 97.00 & 88.95 & 81.31 & 58.25 & 81.38 & 86.55 & 68.83 & 79.37 & 55.91 & 72.66 & \textbf{77.02} & $+0.55$ & 3.88 \\
\bottomrule
\end{tabular}
}
\end{table*}

\section{Reward-Weight Sensitivity}
\label{app:reward_weights}

\noindent\textbf{Setup.}
We perturb each of the three reward weights of CRAFT one at a time and re-train from scratch with the rest of the recipe held fixed. Concretely, we sweep $w_\text{nr} = w_\text{tr}$ (inside $\mathcal{R}_\text{ref}$), $w_\text{c}$ (the spatial consistency reward), and $w_\text{id}$ (the DINO identity reward) at $0.5\times$ and $2\times$ the main-paper values; the remaining weights ($w_\text{t}, w_\text{a}, w_\text{anchor}$) and all training hyperparameters (\secref{app:dataset_size}, \emph{Setup}) are unchanged. Each perturbation is trained for $3{,}000$ optimizer steps and evaluated on XVerseBench under the standard protocol. We do not report a $w_\text{nr}=w_\text{tr}=0.25$ ($0.5\times$) variant because at this weight the $\mathcal{R}_\text{ref}$ signal collapses below the gradient noise of the other rewards, so the run is essentially equivalent to dropping $\mathcal{R}_\text{ref}$ entirely.

\noindent\textbf{Results.}
\tabref{tab:reward_weight_sensitivity} reports per-metric scores. All five perturbations sit within $0.90$ Overall of the baseline (range $75.57$--$77.02$, baseline $76.47$), showing that the recipe is robust to $\pm 2\times$ perturbations of any single weight. Only $w_\text{id}=2.0$ improves Overall, by $+0.55$. The improvement is concentrated in identity columns---Multi ID rises from $61.16$ to $68.83$ ($+7.67$) and Single ID from $84.22$ to $88.95$ ($+4.73$)---while AES drops on both splits ($-2.99$ Single, $-2.25$ Multi) and Single IP loses $2.92$. The other four perturbations either match the baseline or trade a column-level improvement for a comparable loss elsewhere; none reach the baseline's Overall.

\noindent\textbf{Discussion: choice of operating point.}
We treat the sweep as exposing the identity-versus-quality Pareto frontier of CRAFT. Increasing $w_\text{id}$ traces a monotonic trade-off across the three sampled values---Multi ID rises from $57.61 \to 61.16 \to 68.83$ while Multi AES falls from $59.21 \to 58.16 \to 55.91$---confirming that the operating point is a genuine Pareto choice rather than an artifact of any single run.

To pick a balanced point on this frontier rather than the one that maximizes a single aggregate, we rank each configuration on each of the eight raw per-metric columns (Single/Multi $\times$ DPG, ID, IP, AES) and compute a mean rank. Lower is more balanced. The baseline achieves the lowest mean rank ($2.62$), followed by the $w_\text{nr}=w_\text{tr}=2.0$ variant ($3.12$); the configuration that maximizes Overall, $w_\text{id}=2.0$, has mean rank $3.88$ because it is bottom-of-the-table on Single IP, Single AES, and Multi AES while being best on Single ID, Multi ID, and Multi IP. We therefore adopt the baseline as the main configuration: it has the most balanced per-metric profile of the family while preserving aesthetic quality and prompt-aligned IP. A practitioner who prioritizes identity over visual quality---for example, on a face-heavy benchmark or a multi-person editing pipeline---can shift to $w_\text{id}=2.0$ to gain Multi ID at the AES cost reported here. Overall, however, the sweep reshapes the per-metric trade-off (identity vs.\ quality) rather than the aggregate score: no single-weight perturbation either substantially degrades or substantially improves Overall, indicating that CRAFT does not depend on fine-grained reward-weight tuning to remain competitive.

\rev{\noindent\textbf{Sensitivity to non-weight design choices.}
Beyond the reward weights, we vary the three remaining manual choices raised in review---the Gaussian smoothing $\sigma$ applied to the attention maps, the top-$k$ threshold of the attention-derived masks, and the LoRA rank---one at a time, re-training each under the main recipe (\secref{app:dataset_size}) and evaluating on XVerseBench. \tabref{tab:design_sensitivity} varies $\sigma$ and the mask threshold about the main configuration (baseline Overall $76.47$): halving or doubling $\sigma$ ($1.0$/$4.0$) and moving the mask threshold ($k{=}0.3$/$0.7$) shift Overall by at most $0.59$. \tabref{tab:rank_sensitivity} varies the LoRA rank at a fixed training seed: doubling it ($64{\to}128$) changes Overall by only $-0.38$. The reward-locus selection (\secref{app:routing}) is likewise stable across all four ranking criteria. Together with the reward-weight sweep above, no single design knob moves Overall by more than $0.9$, so CRAFT is robust to its design choices rather than dependent on finely tuned settings.}

\begin{table}[t]
\centering\small
\caption{\rev{Sensitivity to non-weight design choices on XVerseBench: Gaussian smoothing $\sigma$ and mask top-$k$ threshold. Each row re-trains the main recipe with a single knob changed. Overall stays within $0.59$ of the baseline.}}
\label{tab:design_sensitivity}
\begin{tabular}{lccc}
\toprule
Variant & Single AVG & Multi AVG & Overall ($\Delta$) \\
\midrule
Baseline ($\sigma{=}2.0$, $k{=}0.5$) & $81.62$ & $71.32$ & $76.47$ \\
$\sigma{=}1.0$ ($0.5\times$) & $81.19$ & $71.03$ & $76.11$ ($-0.36$) \\
$\sigma{=}4.0$ ($2\times$)   & $80.72$ & $71.05$ & $75.88$ ($-0.59$) \\
mask $k{=}0.3$               & $80.69$ & $71.30$ & $76.00$ ($-0.47$) \\
mask $k{=}0.7$               & $81.12$ & $71.13$ & $76.12$ ($-0.34$) \\
\bottomrule
\end{tabular}
\end{table}

\begin{table}[t]
\centering\small
\caption{\rev{Sensitivity to LoRA rank on XVerseBench. Doubling the rank ($64{\to}128$) leaves Overall nearly unchanged.}}
\label{tab:rank_sensitivity}
\begin{tabular}{lccc}
\toprule
LoRA rank & Single AVG & Multi AVG & Overall ($\Delta$) \\
\midrule
$64$ (default) & $81.85$ & $71.30$ & $76.57$ \\
$128$ ($2\times$) & $81.41$ & $70.97$ & $76.19$ ($-0.38$) \\
\bottomrule
\end{tabular}
\end{table}

\section{Limitations}
\label{app:limitations}

CRAFT has \rev{three} main limitations.

\rev{\noindent\textbf{Multi-subject identity separation.}
On the single-subject split, CRAFT achieves its clearest gains; on the multi-subject split, its advantage is smaller---CRAFT still attains the best multi-subject IP, but Multi ID in particular trails the strongest composed-target baseline (MOSAIC; \tabref{tab:xversebench}). This multi-subject identity gap is in part a deliberate identity--quality operating point rather than a hard limit: raising the identity weight $w_\text{id}$ closes much of it at a modest aesthetic cost (\secref{app:reward_weights}), and we report the balanced point in the main paper. Stronger multi-subject identity preservation is left to future work.}

\noindent\textbf{Requires a reference-aware backbone.}
CRAFT presupposes a backbone that natively accepts reference image tokens alongside text and noise: the \emph{Where to look} principle is realized as constraints on the cross-modal attention sub-blocks $\mathbf{A}_{N2R_k}$ and $\mathbf{A}_{N2P_k}$, which exist only when reference tokens are present in the joint sequence. Text-only T2I backbones cannot be fine-tuned with CRAFT without first attaching a reference-conditioning module.

\noindent\textbf{Performance bounded by the initial backbone.}
As a reward-based fine-tuning recipe, CRAFT amplifies the routing behavior already present in the base model rather than rebuilding it from scratch, so the achievable absolute performance is largely determined by the base model's initial state. This is visible in \secref{sec:exp:other-backbone} (\tabref{tab:uno_craft}): applied to UNO~\cite{wu2025uno}---a FLUX.1-dev-based reference adapter---CRAFT improves consistently over the UNO baseline but does not reach the absolute scores of FLUX.2-klein-9B + CRAFT, since UNO has weaker reference-conditioning capacity than FLUX.2-klein-9B. Composing CRAFT on top of UMO~\cite{cheng2025umo} (an already identity-tuned UNO LoRA) partially closes this gap, confirming that a stronger initial state translates into a stronger final model under the same recipe.

\clearpage

\end{document}